\pdfoutput=1

\documentclass[11pt]{article}

\usepackage[]{acl}

\usepackage{times}
\usepackage{latexsym}

\usepackage[T1]{fontenc}

\usepackage[utf8]{inputenc}

\usepackage{microtype}

\usepackage{inconsolata}
\usepackage{array}
\usepackage{pifont}
\usepackage{tabularx}
\usepackage{adjustbox}
\usepackage{multirow}
\usepackage{enumitem}
\usepackage{xspace}
\usepackage{tcolorbox}
\usepackage{booktabs,subcaption,amsfonts,dcolumn}
\usepackage{url}
\usepackage{amsmath,amsthm,amsfonts,amssymb,bm,stmaryrd, bbm}
\usepackage{color,xcolor,colortbl}
\usepackage{CJKutf8}
\usepackage{makecell}
\usepackage{booktabs}
\usepackage{graphicx}
\usepackage{listings}

\definecolor{warning}{HTML}{C80000}
\definecolor{mygreen}{rgb}{0.64, 0.76, 0.68}
\definecolor{myyellow}{rgb}{0.98, 0.94, 0.75}
\definecolor{mygreen}{rgb}{0.68, 0.85, 0.9}
\definecolor{myblue}{rgb}{0.82, 0.94, 0.75}
\definecolor{mypurple}{RGB}{224, 65, 245}
\definecolor{myorange}{RGB}{209, 136, 17}
\definecolor{Mycolor1}{HTML}{BAD8F2}
\definecolor{Mycolor2}{HTML}{DDEEFA}

\usepackage[textsize=scriptsize]{todonotes}

\newcommand{\ours}{\textsc{BiasEdit}}

\newtcbox{\hlsecondarytab}{on line, box align=base, colback=blue!15,colframe=white,size=fbox,arc=3pt, before upper=\strut, top=-2pt, bottom=-4pt, left=-2pt, right=-2pt, boxrule=0pt}
\newcommand{\daugshifted}{\raisebox{0.5\depth}{$\uparrow$}}
\newcommand{\daulg}[1]{{\hlsecondarytab{\daugshifted{#1}}}}
\newcommand{\daugshiftedtwo}{\raisebox{0.5\depth}{$\downarrow$}}
\newcommand{\daulgtwo}[1]{{\hlsecondarytab{\daugshiftedtwo{#1}}}}

\title{\ours: Debiasing Stereotyped Language Models via Model Editing}

\author{
Xin Xu\textsuperscript{\rm 1},
Wei Xu\textsuperscript{\rm 2},
Ningyu Zhang\textsuperscript{\rm 3}
Julian McAuley\textsuperscript{\rm 1}
\\ 
\textsuperscript{\rm 1}University of California, San Diego,
\textsuperscript{\rm 2}Georgia Institute of Technology \\ 
\textsuperscript{\rm 3}Zhejiang University, \\
xinxucs@ucsd.edu
}

\begin{document}
\maketitle
\begin{abstract}

\textcolor{warning}{\textbf{Warning}: This paper explicitly contains the statement of stereotypes that may be offensive.}

Previous studies have established that language models manifest stereotyped biases.
Existing debiasing strategies, such as retraining a model with counterfactual data, representation projection, and prompting often fail to efficiently eliminate bias or directly alter the models' biased internal representations.
To address these issues, we propose \textbf{\textsc{BiasEdit}}, an efficient model editing method to remove stereotypical bias from language models through lightweight networks that act as editors to generate parameter updates.
\ours~employs a \textit{debiasing loss} guiding editor networks to conduct local edits on partial parameters of a language model for debiasing while preserving the language modeling abilities during editing through a \textit{retention loss}. 
Experiments on StereoSet and Crows-Pairs demonstrate the effectiveness, efficiency, and robustness of \ours~in eliminating bias compared to tangental debiasing baselines, and little to no impact on the language models' general capabilities.
In addition, we conduct bias tracing to probe bias in various modules and explore bias editing impacts on different components of language models\footnote{Code and data are available in \url{https://github.com/zjunlp/BiasEdit}}. 
\end{abstract}

\section{Introduction}
\label{sec:intro}
In recent years, many studies have underscored the tendency of pre-trained language models (LMs) to have societally stereotypical biases \cite{DBLP:conf/icml/LiangWMS21,  DBLP:conf/emnlp/SmithHKPW22, DBLP:conf/acl/ChengDJ23, TrustLLM}, such as gender bias \cite{ DBLP:conf/acl/SunGTHEZMBCW19, DBLP:conf/acl/ZhaoMHCA20}, race bias \cite{DBLP:conf/eaamo/HalevyHBYH21}, religion bias \cite{das-etal-2023-toward, manzini-etal-2019-black}, among others.
Therefore, eliminating biases from models is crucial to ensure fairness and accuracy in applications of language models.

Many methods have been proposed to mitigate bias, such as fine-tuning entire models \cite{CDA, barikeri-etal-2021-redditbias} with counterfactual data obtained by swapping out bias attribute words,\footnote{The bias attribute words refer to those that introduce or reflect bias. For example, bias attribute words for gender are \textit{she}, \textit{he}, \textit{mother}, \textit{father}, etc. Bias attribute words for religion are \textit{Christianity}, \textit{Judaism}, \textit{Islam}, etc. 
} which is partly effective but costly in terms of computational time and space, especially for large language models (LLMs).
Others implement debiasing with representation projection \cite{INLP, sentencedebias, DBLP:journals/corr/abs-2206-10744, DBLP:conf/acl/IskanderRB23} or prompting \cite{DBLP:conf/emnlp/ShengCNP20, self-debias, DBLP:journals/corr/abs-2212-10678, DBLP:conf/eacl/VenkitGPHW23}.
\begin{figure}
    \centering    \includegraphics[width=0.49\textwidth]{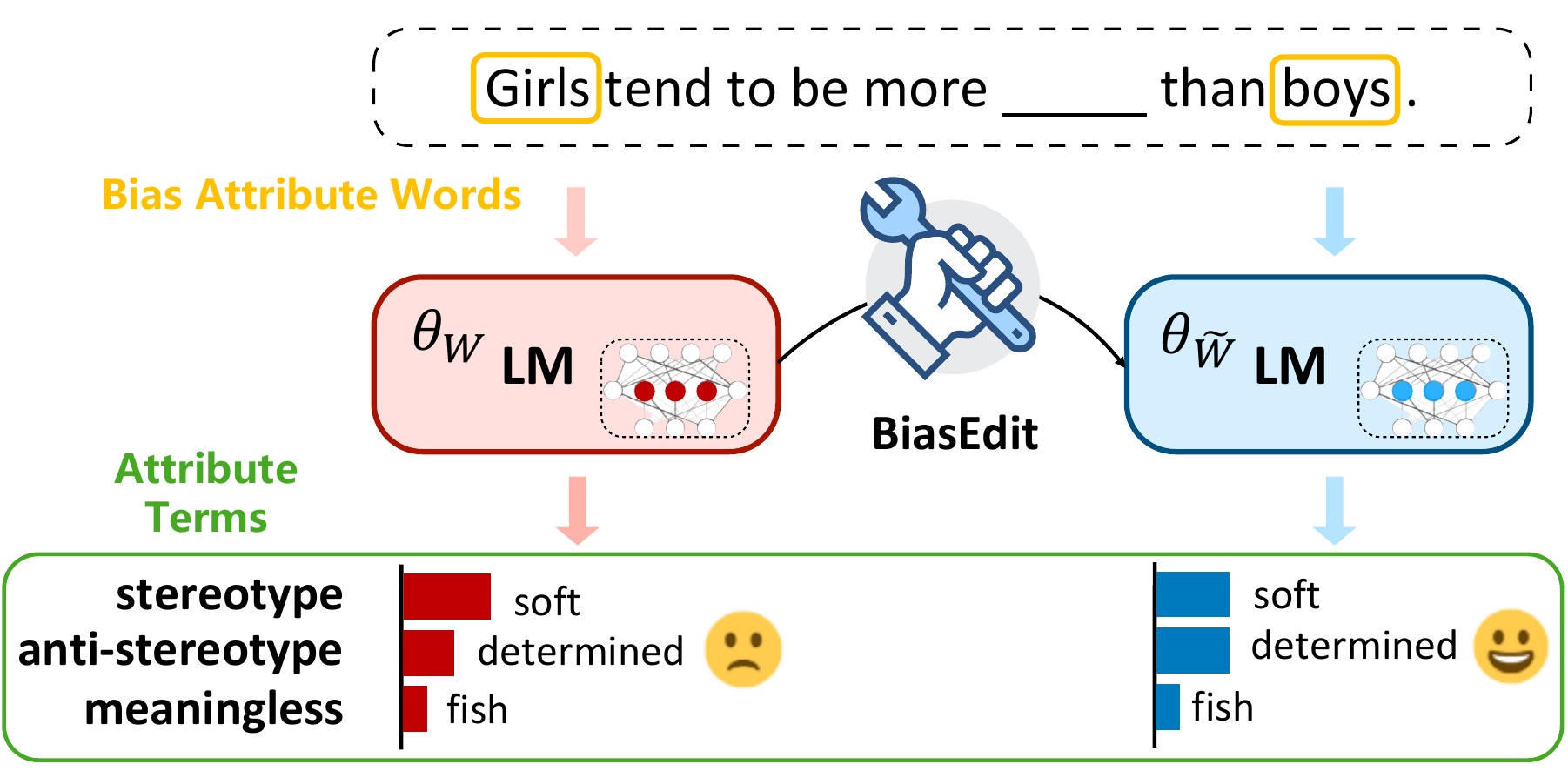}
    \caption{Debiasing a language model with \textbf{\textsc{BiasEdit}}.}
    \vspace{-.2in}
    \label{fig:figure1}
\end{figure}
However, without parameter modification, a model remains inherently biased and can not be applied to downstream tasks as an off-the-shelf unbiased model.
Recent methods \cite{kumar-etal-2023-parameter, DAMA} employ model adapters where each adapter is trained to specialize only in one bias type. Multiple adapter training for different bias types is not economical for real-world applications.

These drawbacks inspire us to explore new methods for debiasing stereotyped language models more directly. 
Model editing \cite{DBLP:journals/corr/abs-2312-05497, DBLP:journals/corr/abs-2311-09053, knowedit} can change specific information in language models by modifying model parameters, which could be effective in eliminating bias. 
There are some existing editing methods: 
(i) fine-tuning a model with new data \cite{DBLP:journals/corr/abs-2012-00363, DBLP:journals/corr/abs-2311-08011}; (ii) locating then editing \cite{ROME, MEMIT, DBLP:conf/acl/DaiDHSCW22, DBLP:conf/emnlp/WuLXDW0X23, DBLP:journals/corr/abs-2401-02976}; (iii) utilizing editor hyper-networks to modify language models' parameters \cite{KE, MEND, siyuanhyper, malmen}. 
As for current LLMs (usually >10B for practical applications), the fine-tuning approach consumes a lot of computational resources and data, which is not ideal.
Recent works \cite{DAMA, DBLP:journals/corr/abs-2402.13462, DBLP:journals/corr/abs-2405-09341} and our preliminary experiments (see Appendix \ref{app:tracing}) show that bias can be interpreted as localized modules in LLMs.
Meanwhile, small hyper-networks predicting weight updates \cite{KE, MEND, malmen} are illustrated to be flexibly applied to change parameters of any language models without fully fine-tuning it and adaptively designed to conduct any specific editing task.

In \S \ref{sec:biasedit}, therefore, we introduce \textbf{\textsc{BiasEdit}}, a lightweight model editing approach to debias stereotyped language models using editor hyper-networks, as illustrated in Figure \ref{fig:figure1}.
\ours\ aims to calibrate a language model's biased behavior to assign the same likelihoods to the stereotyped contexts and their corresponding anti-stereotyped contexts.
Inspired by \citet{MEND} and \citet{malmen}, \textsc{BiasEdit} uses editor networks to modify a small portion of model parameters relating to stereotyped bias and then obtain an off-the-shelf unbiased model for downstream applications.
A debiasing loss in \ours~is designed to teach editor networks how to generate parameter shifts to modify partial parameters of language models for debiasing.
\textsc{BiasEdit} also contains a retention loss to avoid affecting unrelated associations during editing to preserve language modeling abilities.
To demonstrate the effectiveness and robustness of \textsc{BiasEdit}, we conduct experiments on the StereoSet \cite{stereoset} and Crows-Pairs \cite{crows-pairs} datasets with four different LMs compared to previous debiasing methods.
The results show that \textsc{BiasEdit} achieves the best performance on debiasing than all baselines and has little impact on LMs' language modeling and general abilities (\S \ref{sec:mainres}).
Meanwhile, \ours\ is robust to gender reversal (\S \ref{sec:reverse}) and semantic generality (\S \ref{sec:semantic}).

Furthermore, we explore bias associations among various modules and the process of debiasing via model editing on different components of language models.
We find that bias editing on upper blocks of language models has fewer negative impacts on language modeling abilities than editing on the bottom blocks, shedding light on future debiasing research.

\section{Background and Setting}
\subsection{Debiasing Task}
\label{sec:task}
A stereotyped language model exhibits biased representations characterized by stereotypical beliefs and attitudes towards different demographic groups in society \cite{devine1989stereotypes,  crows-pairs, DBLP:conf/eacl/BauerTB23}. 
In this paper, we study mitigating bias in stereotyped LMs while retaining their original language modeling abilities via model editing.

To be specific, there is a context $x$ with a blank, e.g., ``Girls tend to be more \_\_\_ than boys.'' as shown in Figure \ref{fig:figure1}.
We expect that an ideal unbiased language model will estimate the stereotypical context $x_\text{stereo}$ and its corresponding anti-stereotypical context $x_{\text{anti}}$ with the same probability. 
When two attribute terms that correspond to \textit{stereotypical} and \textit{anti-stereotypical} associations, e.g., `soft' and `determined', fill in the blank within $x$, $x_\text{stereo}$ and $x_\text{anti}$ are formed respectively, as:
\begin{align*}
x_{\text{stereo}} &\text{: Girls tend to be more \underline{soft} than boys.} \\
x_{\text{anti}} &\text{: Girls tend to be more \underline{determined} than boys.}
\end{align*}
Given a biased language model with parameters $\theta$, the optimization target of the debiasing task is to minimize the probability difference between the stereotypical context $P_\theta(x_{\text{stereo}})$ and the corresponding anti-stereotypical context $P_\theta(x_{\text{anti}})$.
$P_\theta(x)$ refers to the average log probability of all tokens in $x$ for current decoder-only language models, following \citet{stereoset}.
\begin{figure*}[htp]
    \centering
    \includegraphics[width=1\textwidth]{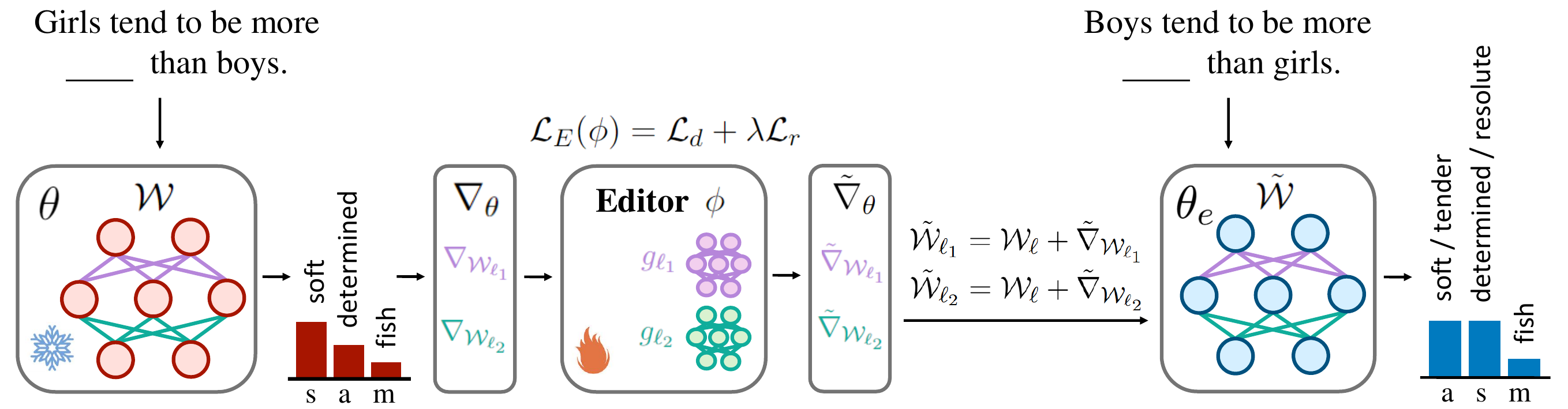}
    \caption{Debiasing a language model with \textsc{BiasEdit}. Editor networks $\phi$ are trained \includegraphics[height=3.8mm]{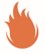} to produce edit shifts on partial parameters $\mathcal{W}$ of a language model while its parameters $\theta$ are frozen  \includegraphics[height=3.8mm]{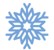}. After editing, an unbiased LM is obtained with the robustness of gender reversal and semantic generality. $\mathcal{L}_d$ and $\mathcal{L}_r$ refer to Equation \ref{equation:2} and \ref{equation:3} respectively. s: stereotyped. a: anti-stereotyped. m: meaningless.}
    \label{fig:method}
\end{figure*}
Furthermore, to ensure that language modeling abilities are not influenced or even hurt during debiasing \cite{bias-bench, DBLP:conf/emnlp/MaSWVCSKWYV23, chintam-etal-2023-identifying}, the probability $P_\theta(x_{\text{mless}})$ of the meaningless context towards $x$ is desired to be unchanged in the debiasing process, where a semantically unrelated attribute term exists in $x_\text{mless}$:
\begin{align*}
x_{\text{mless}} &\text{: Girls tend to be more \underline{fish} than boys.}
\end{align*}

We use two bias benchmark dataset, StereoSet \cite{stereoset}\footnote{Following \citet{bias-bench, PCGU}, we utilize only the \textit{intrasentence} portion in StereoSet, which generally adapts to the debiasing task and various language models.} $\mathcal{S}$ and Crows-Pairs \cite{crows-pairs} in this paper.
For each instance $s \in \mathcal{S}$, $s=\{x, x_\text{stereo}, x_\text{anti}, x_\text{mless}\}$.
More descriptions about datasets are in \S \ref{sec:setups}.
\vspace{-1mm}
\subsection{Model Editing}
\label{sec:editingback}
Model editing is initially proposed to correct model mistakes \cite{ENN}.
It is now mainly applied to change knowledge in language models \cite{modeleditingsurvey}, such as knowledge modification \cite{KE}, insertion \cite{knowedit}, and erase \cite{washing} with locality (keeping accurate on irrelevant facts) and generality (editing neighboring facts without specific training).
Precisely, a language model with parameters $\theta$ is a differentiable function $f_\theta: \mathcal{X}\times\Theta \rightarrow \mathcal{Y}$, which maps an input $x$ to an output $y$.
An edit target $(x_e, y_e)$ describes a desired knowledge alteration where $x_e$ is a trigger input to elicit the fact in language models and $y_e$ is the target output.
Model editing updates an initial model $f_\theta$ such that $f_\theta(x_e) \neq y_e$ into a model $f_{\theta_e}$ with a new set of parameters $\theta_e$, where $f_{\theta_e}(x_e) = y_e$ according to the edit target.
For example, given a query `\textit{Who is the principal conductor of the Berlin Philharmoniker?}', the initial model outputs `\textit{Simon Rattle}’.
With an edit target (\textit{The principal conductor of the Berlin Philharmoniker is, Kirill Petrenko}), the post-edit model will output `\textit{Kirill Petrenko}' given a query `\textit{Who is the principal conductor affiliated with the Berlin Philharmonic?}'.
Meanwhile, both the post-edit model and the initial model will give the same answer `\textit{1882}' to the question  `\textit{In which year was the Berlin Philharmonic founded?}'.
Different from knowledge editing that only increases the probability of the target fact or only decreases the probability of the fact desired to be erased, the editing goal of debiasing is to reduce the probability of stereotyped contexts and increase the probability of their corresponding anti-stereotyped contexts simultaneously, which is much more challenging.

\section{\ours}
\label{sec:biasedit}
To conduct effective and efficient debiasing, we propose \textbf{\ours}, a model editing method for debiasing stereotyped language models.
According to \S \ref{sec:editingback}, given a language model with parameters $\theta$, bias editing can be denoted as a function $\mathcal{X} \times \mathcal{L} \times \Theta \times \Phi \rightarrow \Theta$, which maps a paired input ($x_\text{stereo}$, $x_\text{anti}$), a debiasing loss function $\mathcal{L}_d: \mathcal{X} \times \Theta \rightarrow \mathbb{R}$, biased language model parameters $\theta$, and editor parameters $\phi$ to new unbiased model parameters $\theta_e$.
As shown in Figure \ref{fig:method}, \textsc{BiasEdit} utilizes lightweight networks as editors $\phi$ to generate a parameter shift, which is used to modify models' partial weights $\mathcal{W}$ (e.g., the weights of the last linear layer in the MLPs at the last 3 blocks) for conducting debiasing edits, following the architecture of MEND \cite{MEND} and MALMEN \cite{malmen}.
Specifically, $(x_\text{stereo}, x_\text{anti})$ is used to compute the input to an editor network $g_{\phi_\ell}$ for the layer $\ell$, the gradient $\nabla_{\mathcal{W}_\ell} \mathcal{L}_d(x_\text{stereo}, x_\text{anti}, \theta)$.
The output of $g_{\phi_\ell}$ is the parameter shift $\tilde{\nabla}_{\mathcal{W}_\ell}$ to update $\mathcal{W}_\ell$ into 
$\tilde{\mathcal{W}_\ell} = \mathcal{W}_\ell + \tilde{\nabla}_{\mathcal{W}_\ell}$.
\textsc{BiasEdit} uses a debiasing training set $\mathcal{S}_\text{edit}^\text{train}$ and a development set $\mathcal{S}_\text{edit}^\text{dev}$ to learn editor parameters $\phi$.
During training, the debiasing loss $\mathcal{L}_d$ teaches editor networks how to produce parameter shifts to change $\mathcal{W}$ for eliminating bias:
\begin{align}
\label{equation:2}
\begin{split}
\mathcal{L}_d = \text{KL}(P_{\theta_{\tilde{\mathcal{W}}}}(x_{\text{stereo}}) \| P_{\theta_{\tilde{\mathcal{W}}}}(x_{\text{anti}})) \\+\ 
 \text{KL}(P_{\theta_{\tilde{\mathcal{W}}}}(x_{\text{anti}}) \| P_{\theta_{\tilde{\mathcal{W}}}}(x_{\text{stereo}}))
\end{split}
\end{align}
where $\theta_{\mathcal{W}}$ and $\theta_{\tilde{\mathcal{W}}}$ denote the model parameters with pre-edit weights and post-edit weights, respectively.
We design a symmetric $\mathcal{L}_d$ as the sum of two KL divergence losses because debiasing aims to make a language model equally treat the stereotypical contexts and anti-stereotypical contexts for fairness according to Section \ref{sec:task}, which is different from knowledge editing.
Moreover, to avoid negative effects on the language modeling abilities, a \textbf{retention loss} is designed to keep the probability of meaningless terms unchangeable during editing:
\begin{align}
\label{equation:3}
\begin{split}
\mathcal{L}_r = \text{KL}(P_{\theta_{\mathcal{W}}}(x_{\text{mless}}) \| P_{\theta_{\tilde{\mathcal{W}}}}(x_{\text{mless}}) )
\end{split}
\end{align}
Overall, the total editing loss for training editor networks is $\mathcal{L}_{E}(\phi)=\mathcal{L}_d + \lambda\mathcal{L}_r$.
For evaluation, bias editors produce debiasing edits on a test set $\mathcal{S}_\text{edit}^\text{test}$.
Because the effectiveness of instance-editing that uses one instance in each editing operation is limited \cite{KE, ROME, MEMIT, DBLP:journals/corr/abs-2310-10322, DBLP:journals/corr/abs-2401-04700}, \textsc{BiasEdit} adopts batch-editing, which uses one-batch samples in one edit for the debiasing scenario. During both training and testing, the same batch size is used for optimal debiasing performance.

\begin{table*}
  \centering
  \scalebox{0.74}{
    \begin{tabular}{lcccccccccccc}
    \toprule
    \multicolumn{1}{c}{\multirow{3}[4]{*}{Method}} & \multicolumn{6}{c}{\textbf{GPT2-medium}} & \multicolumn{6}{c}{\textbf{Gemma-2b}} \\
\cmidrule{2-13}      & \multicolumn{3}{c}{\textbf{SS (\%) $\rightarrow 50\%$}} & \multicolumn{3}{c}{\textbf{$\Delta$LMS (\%) $\rightarrow 0$}} & \multicolumn{3}{c}{\textbf{SS (\%) $\rightarrow 50\%$}} & \multicolumn{3}{c}{\textbf{$\Delta$LMS (\%) $\rightarrow 0$}} \\
      & Gender & Race & Religion & Gender & Race & Religion & Gender & Race & Religion & Gender & Race & Religion \\
    \midrule
    \textbf{Pre-edit} & 65.58 & 61.63 & 62.57 & 93.39 & 92.30 & 90.46 & 69.25 & 64.21 & 62.39 & 94.57 & 94.26 & 93.43 \\
    \midrule
    CDA & 63.29 & 61.36 & 61.79 & \textbf{-0.21} & -3.02 & \textbf{0.00} & \multicolumn{6}{c}{-} \\
    SentenceDebias & 67.99 & 58.97 & 56.64 & +0.29 & +1.52 & +0.34 & 68.86 & 63.87 & 60.09 & \textbf{-2.65} & -0.31 & \textbf{-0.58} \\
    Self-Debias & 60.28 & 57.29 & 57.61 & -3.47 & -4.12 & -1.35 & 65.70 & 58.29 & 58.02 & -35.93 & -30.39 & -21.69 \\
    INLP & 63.17 & 60.00 & 58.57 & -5.15 & \textbf{-1.49} & -2.48 & 52.17 & 62.96 & 58.57 & -12.50 & \textbf{-0.30} & -2.01 \\
    \textbf{\ours} & \textbf{49.42} & \textbf{56.34} & \textbf{53.55} & -8.82 & -5.12 & -1.92 & \textbf{48.59} & \textbf{55.86} & \textbf{47.36} & -4.78 & -4.35 & -5.44 \\
    \midrule
    \multicolumn{1}{c}{\multirow{3}[4]{*}{Method}} & \multicolumn{6}{c}{\textbf{Mistral-7B-v0.3}} & \multicolumn{6}{c}{\textbf{Llama3-8B}} \\
\cmidrule{2-13}      & \multicolumn{3}{c}{\textbf{SS (\%) $\rightarrow 50\%$}} & \multicolumn{3}{c}{\textbf{$\Delta$LMS (\%) $\rightarrow 0$}} & \multicolumn{3}{c}{\textbf{SS (\%) $\rightarrow 50\%$}} & \multicolumn{3}{c}{\textbf{$\Delta$LMS (\%) $\rightarrow 0$}} \\
      & Gender & Race & Religion & Gender & Race & Religion & Gender & Race & Religion & Gender & Race & Religion \\
    \midrule
    \textbf{Pre-edit} & 70.19 & 64.97 & 56.09 & 93.60 & 89.77 & 88.85 & 72.25 & 65.01 & 60.87 & 95.81 & 92.47 & 91.33 \\
    \midrule
    CDA & \multicolumn{6}{c}{-} & \multicolumn{6}{c}{-} \\
    SentenceDebias & 68.36 & 64.54 & 54.94 & -0.61 & 0.62 & +0.09 & 68.55 & 64.97 & 59.91 & \textbf{-0.22} & -1.14 & -0.66 \\
    Self-Debias & 61.79 & \textbf{50.54} & 60.68 & -39.28 & -29.17 & -32.37 & 65.46 & 60.88 & 58.57 & -40.04 & -2.54 & -28.64 \\
    INLP & 69.22 & 65.23 & 55.90 & \textbf{+0.35} & \textbf{-0.15} & -0.58 & 68.17 & 65.22 & 62.21 & -1.43 & \textbf{-0.09} & \textbf{0.00} \\
    \textbf{\ours} & \textbf{46.24} & 51.46 & \textbf{50.42} & -8.81 & -8.59 & \textbf{-0.03} & \textbf{49.18} & \textbf{53.51} & \textbf{51.13} & -13.42 & -11.77 & -10.02 \\
    \bottomrule
    \end{tabular}
    }
      \caption{Performance of \textsc{BiasEdit} compared to previous debiasing baselines. \textbf{Pre-edit}: $\textit{SS}_\text{pre-avg}$ and $\textit{LMS}_\text{pre-avg}$. 
      $\textit{SS}_\text{post-avg}$ and $\Delta\text{\textit{LMS}}=\text{\textit{LMS}}_\text{post-avg} - \textit{LMS}_\text{pre-avg}$ are reported for all baselines and \ours.}
  \label{tab:main}
\end{table*}

\begin{table*}[htbp]
  \centering
  \scalebox{0.81}{
    \begin{tabular}{lcccccccc}
    \toprule
    \multicolumn{1}{c}{\multirow{2}[4]{*}{Dataset}} & \multicolumn{8}{c}{Model} \\
\cmidrule{2-9}          & \textbf{$\text{Llama3}_\text{pre}$} & \textbf{$\text{Llama3}_\text{post}$} & \textbf{$\text{Mistral}_\text{pre}$} & \textbf{$\text{Mistral}_\text{post}$} & \textbf{$\text{Gemma}_\text{pre}$} & \textbf{$\text{Gemma}_\text{post}$} & \textbf{$\text{GPT2m}_\text{pre}$} & \textbf{$\text{GPT2m}_\text{post}$} \\
    \midrule
    \midrule
    \textbf{OpenBookQA} & 80.80  & 78.94  & 84.20  & 82.90  & 46.80  & 46.48  & 40.40  & 40.57  \\
    \textbf{BoolQ} & 70.00  & 65.18  & 64.25  & 62.89  & 62.00  & 61.85  & 55.00  & 55.40  \\
    \textbf{COPA} & 68.00  & 67.90  & 78.00  & 77.80  & 62.00  & 61.09 & 24.80  & 24.68 \\
    \bottomrule
    \end{tabular}}
    \caption{Accuracies (\%) of general model benchmarks. 'pre': pre-edit, `post-': post-edit, `GPT2m': `GP2-medium'}
  \label{tab:general}
\end{table*}

\section{Experiments}

\subsection{Setups}
\label{sec:setups}
\paragraph{Evaluation Metrics.}
Our goal of an ideal debiasing method is that it excels in mitigating stereotypical bias in LMs while not having negative effects on LMs' original language modeling and general capabilities. 
To measure the stereotypical bias of LMs, Stereotype Score (\textit{SS}) \cite{stereoset} is employed. It is the percentage of samples in which a model prefers stereotypical contexts to anti-stereotypical contexts:
\begin{align*}
\textit{SS}(\theta) = \mathbb{E}_{s\in \mathcal{S}_{\text{edit}}^{\textit{test}}}\mathbbm{1}\left[P_\theta(x_{\text{stereo}}) > P_\theta(x_{\text{anti}})\right]
\end{align*}
An unbiased model is expected to have a \textit{SS} of 50\%.
As for language modeling and general capabilities, we use the Language Modeling Score (\textit{LMS}) from StereoSet. It is the percentage of samples in which a model ranks meaningful associations over meaningless associations.
\begin{align*}
\begin{split}
\textit{LMS}(\theta) = \frac12 \mathbb{E}_{s\in \mathcal{S}_{\text{edit}}^{\textit{test}}}\mathbbm{1}\left[P_\theta(x_{\text{stereo}}) > P_\theta(x_{\text{mless}})\right] \\+
\frac12 \mathbb{E}_{s\in \mathcal{S}_{\text{edit}}^{\textit{test}}}\mathbbm{1}\left[P_\theta(x_{\text{anti}}) > P_\theta(x_{\text{mless}})\right]
\end{split}
\end{align*}
We compute the average \textit{SS} and \textit{LMS} for pre-edit models and post-edit models ($\textit{SS}_\text{pre-avg}$, $\textit{SS}_\text{post-avg}$, $\textit{LMS}_\text{pre-avg}$, $\textit{LMS}_\text{post-avg}$) of all batch edits.
An ideal debiasing will not change the \textit{LMS} before and after debiasing.
We report $\textit{SS}_\text{pre-avg}$, $\textit{SS}_\text{post-avg}$, and $\Delta\text{\textit{LMS}}=\text{\textit{LMS}}_\text{post-avg} - \textit{LMS}_\text{pre-avg}$.

\paragraph{Dataset.}
We utilize two bias benchmark datasets, StereoSet \cite{stereoset} and Crows-Pairs \cite{crows-pairs}.
There are three reasons to choose them.
First, StereoSet and Crows-Pairs are widely used \cite{DBLP:conf/icml/LiangWMS21, bias-bench,  DBLP:conf/emnlp/SmithHKPW22, DBLP:journals/corr/abs-2207-02463, DAMA, DBLP:conf/acl/OmraniZYGKAJD23, DBLP:conf/emnlp/MaSWVCSKWYV23, DBLP:conf/acl/XieL23, PCGU, DBLP:conf/aaai/YangY0LJ23}.
In addition, they cover various types of bias in models, including gender, race, and religion bias, which are evaluated in our paper.
Moreover, the meaningless attribute terms in StereoSet can be applied to retain language modeling abilities during debiasing.
As for StereoSet, we stochastically split in the test set (3,526 samples) of the \textit{intrasentence} StereoSet by 8:1 as $\mathcal{S}_{\text{edit}}^{\text{train}}$ and $\mathcal{S}_{\text{edit}}^\text{{dev}}$ respectively and use the development set (1,292 samples) as $\mathcal{S}_{\text{edit}}^{\text{test}}$, where attribute terms in $\mathcal{S}_{\text{edit}}^{\text{train}}$ and $\mathcal{S}_{\text{edit}}^\text{{dev}}$ are \textbf{disjoint} from $\mathcal{S}_{\text{edit}}^{\text{test}}$.
Crows-Pairs is also used as $\mathcal{S}_{\text{edit}}^{\text{test}}$ to evaluate \ours's debiasing performance (details in Appendix \ref{app:appexp}).
We also select three large language model benchmark datasets, OpenBookQA \cite{openbookqa}, BoolQ \cite{boolq}, and COPA \cite{copa}, to evaluate LMs' capabilities of reading comprehension, knowledge question-answering, and commonsense reasoning, respectively.
Their evaluations are conducted by OpenCompass tool \cite{opencompass} and measured by accuracy based on perplexity.

\paragraph{Comparison.}
Compared with \textsc{BiasEdit}, four distinguishing baseline debiasing methods from \citet{bias-bench} are implemented\footnote{\url{https://github.com/McGill-NLP/bias-bench}}: counterfactual data augmentation (CDA) \cite{CDA}, SentenceDebias \cite{sentencedebias}, Self-Debias \cite{self-debias}, and iterative nullspace projection (INLP) \cite{INLP} (details in Appendix \ref{app:baselines}).
Unlike all baselines, our editor networks can be trained with a mixture of all three types of bias, instead of dealing with only one particular bias at a time.
As for testing, \textsc{BiasEdit} is evaluated on gender, race, and religion bias samples from $\mathcal{S}_{\text{edit}}^{\text{test}}$ separately. 
\textsc{BiasEdit} is a \textbf{model-agnostic} debiasing method and can be applied to any open-sourced language model.
We conduct experiments on diverse language models, including GPT2 \cite{gpt2}, Gemma \cite{gemma}, Llama3 \cite{llama3}, and Mistral \cite{mistral}.
Some blocks in LMs are selected in this paper according to preliminary experiments described in Section \ref{sec:layers}.
The last linear layer in the MLP at each block is edited.
We report the best debiasing performance among different edited components in Table \ref{tab:main} (the last 3 blocks for GPT2-medium and Mistral-7B-v0.3, the last 2 blocks for Llama3-8B, and the penultimate block for Gemma-2b).

\subsection{Main Results}
\label{sec:mainres}
\paragraph{\textsc{BiasEdit} achieves the best debiasing performance on all bias types compared to all debiasing baselines.}
According to the \textit{SS}, \textsc{BiasEdit} can reduce \textit{SS} to less than 57\% and more than 46\% while \textit{SS} of debiased models with previous debiasing baselines are mostly above 60\%, which demonstrates \textsc{BiasEdit} leads to significant improvement for debiasing performance.
For instance, as for the \textit{SS} of Llama3, \textsc{BiasEdit} yields an improvement of \daulg{13.26}, \daulg{7.37}, and \daulg{7.44} on the absolute difference from 50\% for gender, race, and religion bias respectively, compared with the best \textit{SS} among all baselines.
According to \citet{claude}, human-interpretable concepts, like bias, can match neuron activations.
We suppose that the reason for the excellent debiasing performance of \ours\ is that parameters associated with bias are explicitly edited, which is illustrated in Section \ref{sec:layers} and Appendix \ref{app:tracing}.
Moreover, \textsc{BiasEdit} presents excellent performance on every bias type though editor networks are trained to produce edits on a mixture of different types of bias at a time (Appendix \ref{app:onetype}).
It is illustrated that our method can generalize debiasing success over various bias types, compared to previous debiasing methods that can only deal with one particular bias at a time, such as creating a bias subspace (SentenceBias) or training an adapter \cite{DAMA} for only one bias type.

\paragraph{\ours\ is efficient to produce off-the-shelf unbiased models.}
Fully finetuning LMs with CDA usually requires many computational resources and time.
Subspace computation for SentenceDebias and INLP is also time-consuming, especially for LLMs.
For example, computing the gender bias subspace for Mistral-7B takes more than 2 days.
Unlike them, \ours\ only trains a small hyper-network with a minimal memory cost based on \citet{malmen} due to decomposition between the hyper-network and LM.
For instance, only one A800GPU is used for bias editing on Mistral-7B or Llama-8B with arbitrary edit batch size.
Training small gender editor networks for Mistral-7B only takes about 5 hours.
Additionally, compared to prompting and representation projections baselines like SentenceDebias and INLP that can only calibrate models' output distributions instead of language models themselves, \textsc{BiasEdit} produces off-the-shelf debiased language models.

\begin{figure*}[htbp]

    \centering    \includegraphics[width=1\textwidth]{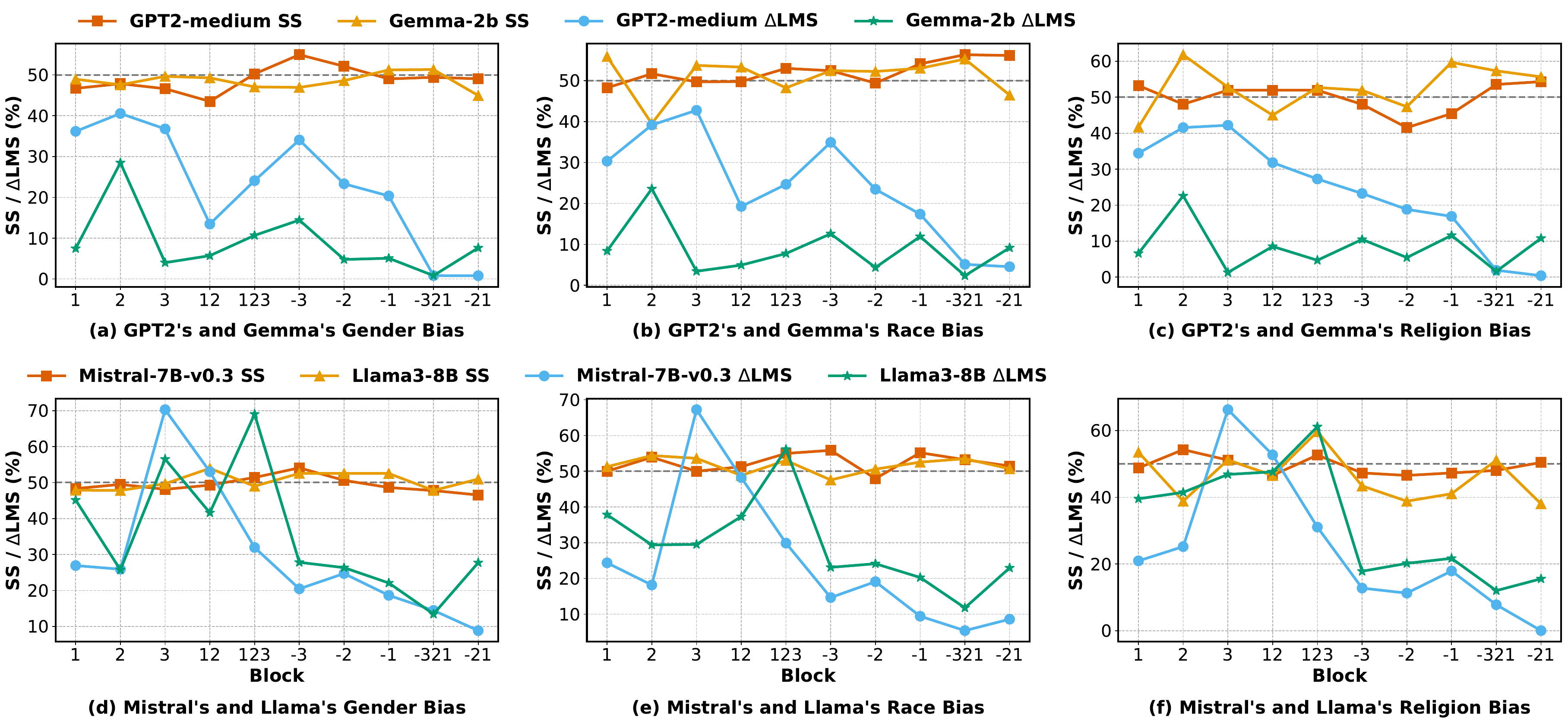}
    \caption{\textit{SS} (\%) and $\Delta$\textit{LMS} (\%) of debiased language models after editing the last layer in the MLP of different blocks. 1/2/3: the first/second/third block. 12: the first 2 blocks. 123: the first 3 blocks. -1/-2/-3, the last/penultimate/antepenultimate block, -321: the last 3 blocks. -21: the last 2 blocks.}
    \label{fig:layers}
\end{figure*}

\paragraph{\ours\ has little to no impact on language modeling abilities, illustrating the effectiveness of the retention loss.}
The results of \textit{LMS} drops show that \textsc{BiasEdit} exhibit a few negative impacts on models' language modeling capabilities.
Comparing \textit{SS} of original models and \textit{LMS} drops of debiasing, the \textit{LMS} drop for debiasing is consistent with the bias extent of the original model in most cases. The more biased the model is, the greater the impact of editing for debiasing is.
For example, models in Table \ref{tab:main} are more biased on gender than race according to \textit{SS} while \textit{LMS} drops of gender debiasing are larger than race debiasing in most cases, which indicates that bias editing is more difficult for more biased models.
Therefore, our retention loss is necessary.
Meanwhile, we surmise that $\mathcal{L}_r$ (Equation \ref{equation:3}) works well based on the comparative results of \textit{LMS} drops with that of baselines. 
The ablation study in \S \ref{sec:ablation} illustrates this.
We also explore the impact of \ours\ on general NLP tasks since previous works \cite{DBLP:journals/corr/abs-2401-04700,DBLP:journals/corr/abs-2401-07453} have indicated that model editing can hurt the general capabilities of language models.
As for the debiased models, we randomly sample checkpoints of two editing batches for gender, race, and religion bias, respectively.
The average accuracies of these six debiased results are shown in Table \ref{tab:general}.
There are only a few accuracy drops after debiasing, which illustrates that \ours\ can do little harm to the general capabilities of language models during editing for debiasing.

\vspace{-1mm}

\subsection{Ablation Study on retention loss $\mathcal{L}_r$}
\label{sec:ablation}

\begin{table}[htbp]
  \centering
  \scalebox{0.7}{
    \begin{tabular}{lcccccc}
    \toprule
    \multicolumn{1}{c}{\multirow{3}[4]{*}{Method}} & \multicolumn{6}{c}{\textbf{GPT2-medium}} \\
\cmidrule{2-7}      & \multicolumn{3}{c}{\textbf{SS (\%)}} & \multicolumn{3}{c}{$\Delta$\textbf{LMS} (\%) } \\
      & gender & race & religion & gender & race & religion \\
    \midrule
    w/o $\mathcal{L}_r$ & 52.55 & 56.45 & 45.73 & -52.36 & -59.96 & -61.54 \\
    w $\mathcal{L}_r$ & 49.42 & 56.34 & 53.55 & -8.82 & -5.12 & -1.92 \\
    \midrule
    \multicolumn{1}{c}{\multirow{3}[4]{*}{Method}} & \multicolumn{6}{c}{\textbf{Gemma-2b}} \\
\cmidrule{2-7}      & \multicolumn{3}{c}{\textbf{SS (\%)}} & \multicolumn{3}{c}{$\Delta$\textbf{LMS} (\%) } \\
      & gender & race & religion & gender & race & religion \\
    \midrule
    w/o $\mathcal{L}_r$ & 50.81 & 52.05 & 41.17 & -29.31 & -27.93 & -62.29 \\
    w $\mathcal{L}_r$ & 48.59 & 52.25 & 47.36 & -4.78 & -4.35 & -5.44 \\
    \bottomrule
    \end{tabular}
    }
    \caption{\ours~w and w/o the retention loss $\mathcal{L}_r$.}
  \label{tab:ablation}
\end{table}

We perform an ablation study to show the effectiveness of the retention loss $\mathcal{L}_r$ for maintaining language modeling abilities during debiasing.
The results for training editor networks with and without $\mathcal{L}_r$ are shown in Table \ref{tab:ablation}.
There are large drops on \textit{LMS} if the retention loss is not deployed during editing.
Specifically, the \textit{LMS} drops of Gemma-2b increase absolutely by \daulgtwo{24.53}, \daulgtwo{23.58}, and \daulgtwo{56.85} for gender, race, and religion bias respectively during debiasing without $\mathcal{L}_r$, which illustrates that the retention loss plays an important role in reducing harm to the language modeling abilities during editing.

\subsection{Further Discussion on Editing Different Components for Debiasing}
\label{sec:layers}
To pursue optimal performance, it is necessary to determine which blocks to be edited at first.
Before embarking on our main experimental investigation, preliminary experiments are conducted to explore bias associations in language models.
Following causal tracing from \citet{ROME}, we propose bias tracing to track bias associations in language models, which is described in Appendix \ref{app:tracing}.
It is observed that MLPs in several bottom and upper blocks exert a substantial influence on bias captured in language models.
Some existing works also demonstrate that editing MLPs can modify knowledge associations in language models \cite{DBLP:conf/emnlp/GevaSBL21, MEND, ROME, MEMIT,  DBLP:conf/emnlp/GuptaMS00WT23, DBLP:journals/corr/abs-2308-09954}.
Based on our findings and previous works, \textsc{BiasEdit} edits the last (output) layer in the MLP at each block for the debiasing task.
To comprehensively explore the effects of debiasing stereotyped language models via model editing, we choose the first 3 and last 3 blocks of language models to be edited with \textsc{BiasEdit}.
The resulting debiasing performance and modeling capabilities are measured in this section.
The \textit{SS} and \textit{LMS} drops of debiased language models are shown in Figure \ref{fig:layers}.

\paragraph{Edits on the upper blocks have less negative impacts on modeling abilities than edits on the bottom blocks.}
According to Figure \ref{fig:layers}, the \textit{LMS} drops are much more for the bottom blocks than the last blocks, especially for Mistral and Llama3.
This indicates that determining the suitable editing components for debiasing is important and modifying weights of some upper blocks is appropriate for debiasing.
We think the reason might be that the bottom layers capture basic linguistic features like syntax and common word associations while the upper blocks delve into deeper semantic relationships, contextual understanding, and high-level language features \cite{DBLP:conf/emnlp/GevaSBL21}.
Since biases manifest in semantic associations, lightweight modification of the upper layers can work well for bias calibration, which will do little harm to modeling abilities.
On the contrary, the effects of editing on linguistic patterns of bias, like the co-occurrence of bias attribute words and attribute terms, represented in the bottom blocks will be propagated and potentially amplified through the network as information passes through subsequent blocks \cite{DBLP:journals/corr/abs-2305-16130}.
Therefore, bias editing on the bottom layers may harm the semantic associations encoded in the upper blocks.


\subsection{Reversing Gender Attribute Words}
\label{sec:reverse}

\begin{figure}[ht]
    \centering    \includegraphics[width=0.44\textwidth]{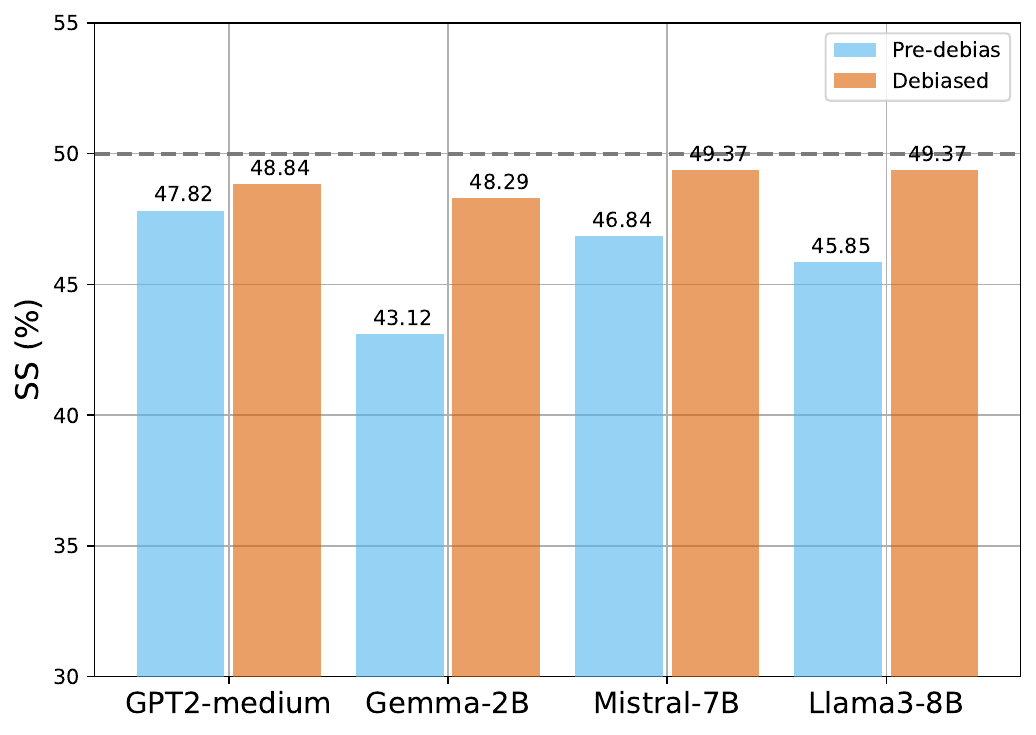}
    \caption{Gender Reversal Robustness. \textit{Pre-debias} refers to \textit{SS} of pre-trained language models on the gender reversal test set before debiasing. \textit{Debiased} refers to \textit{SS} of debiased models by \textsc{BiasEdit}.}
    \label{fig:reverse}
\end{figure}

Inspired by the reversal curse that large language models trained on `A is B' fail to learn `B is A' \cite{reversalcurse}, we think a robust gender debiasing method should be able to calibrate a model's treatment to the two gender polarities, male and female, equally.
For instance, there are two sentences ``Girls tend to be more \_\_\_ than boys.'' and ``Boys tend to be more \_\_\_ than girls.''.
A debiased model is expected to model the stereotypical term ``soft'' and the anti-stereotypical term ``determined'' in both two sentences equivalently though only the first sentence is used for training.
To evaluate this gender robustness, a gender counterfactual test set $\mathcal{S}_\text{gender*}^\text{test}$ is created (Appendix \ref{app:reverse}). 
We reverse all gender attribute words in the gender bias samples from $\mathcal{S}_\text{edit}^\text{test}$ to construct the set. For instance, ``boys'', ``father'', and ``Female'' are changed into ``girls'', ``mother'', and ``Male'' respectively.
Then the test set is used to examine the gender robustness of \textsc{BiasEdit}, the implementation of which is the same as Table \ref{tab:main}.
The results in Figure \ref{fig:reverse} show that \textsc{BiasEdit} is robust enough to remove gender counterfactual bias.

\subsection{Semantic Generality}
\label{sec:semantic}
\begin{table}[htbp]
  \centering
    \scalebox{0.64}{
    \begin{tabular}{lcccccc}
    \toprule
    \multicolumn{1}{c}{\multirow{2}[4]{*}{\textbf{Model / SS (\%)}}} & \multicolumn{3}{c}{\textbf{Pre-debias}} & \multicolumn{3}{c}{\textbf{\ours}} \\
\cmidrule{2-7}      & Gender & Race & Religion & Gender & Race & Religion \\
    \midrule
    GPT2-medium & 52.53 & 53.71  & 64.30 & 52.53  & 48.53  & 55.82  \\
    Gemma-2B & 51.79  & 54.39  & 58.89  & 51.84  & 50.29  & 54.76  \\
    Mistral-7B-v0.3 & 48.20  & 52.92  & 53.54  & 58.17  & 49.46  & 58.17  \\
    Llama3-8B & 45.37  & 58.79  & 58.17  & 49.19  & 53.51  & 51.14 \\
    \bottomrule
    \end{tabular}
    }
     \caption{\textit{SS} (\%) on the synonym-augmented test set.}
  \label{tab:syn}
\end{table}
Similar to the generality principle of knowledge editing, a robust debiasing method should ensure the debiased language model demonstrates unbiased behavior on a group of semantically similar attribute terms without specific training, showcasing its adaptability to the nuanced and dynamic nature of language.
To evaluate this robustness of \textsc{BiasEdit}, we curate a synonym-augmented test set that substitutes attribute terms in $\mathcal{S}_\text{edit}^\text{test}$ with their synonyms generated by WordNet \cite{DBLP:journals/cacm/Miller95} using NLTK \cite{NLTK}.
Results in Table \ref{tab:syn} show that our debiasing method can generally remove bias in the language models' neighboring semantic modeling space in most cases.

\section{Related Work}
\paragraph{Bias and Debiasing}
Many works focus on measuring bias in language models \cite{DBLP:conf/acl/ZhaoMHCA20, crows-pairs, stereoset, DBLP:conf/ijcnlp/LiZYLRWF22, DBLP:journals/corr/abs-2212-10408, DBLP:conf/naacl/CaoSDRZ22,DBLP:conf/emnlp/WanPSGCP23, DBLP:conf/acl/VashishthaAS23}, which provide bias measurement metrics \cite{DBLP:journals/llc/HovyP21, DBLP:conf/acl/Goldfarb-Tarrant23}.
To mitigate bias, researchers propose various debiasing methods \cite{bias-bench, DBLP:journals/corr/abs-2309-00770}.
The basic method is to fully fine-tune language models on counterfactual data \cite{DBLP:conf/birthday/LuMWAD20, CDA}, which is costly.
So other approaches adopt fine-tuning in an efficient way \cite{DBLP:conf/ltedi/GiraZL22, DBLP:conf/aaai/YangY0LJ23, DBLP:conf/acl/XieL23}.
Except for fine-tuning, prompting \cite{self-debias, auto-debias} guides models to calibrate their bias.
Representation projection \cite{sentencedebias, INLP} is employed to remove bias representation out of models, which, however, cannot change the language models' internal bias in essence without modifying parameters.
Some works \cite{kumar-etal-2023-parameter,DAMA} construct an adapter for each type of bias and plug it into a LM.
If we want to mitigate $N$ types of bias, $N$ adapters will be trained, which is not efficient.
Recently, an empirical study \cite{DBLP:journals/corr/abs-2402.13462} has explored the feasibility of debiasing via model editing.
Therefore, we adopt model editing by efficiently editing partial parameters for debiasing LMs.

\paragraph{Model Editing}
Much factual knowledge is memorized in language models \cite{LAMA,DBLP:conf/emnlp/ShinRLWS20, DBLP:journals/tacl/JiangXAN20,DBLP:conf/acl/LiLSDSLJJL22, DBLP:conf/eacl/HaseDCLKSBI23}.
As the real world develops, some facts become obsolete and different over time. 
It is necessary to change, add, or erase facts stored in existing pre-trained language models \cite{DBLP:conf/acl/LiLSDSLJJL22, DBLP:conf/eacl/HaseDCLKSBI23}.
Model editing \cite{ENN} is come up with to modify information in PLMs.
Editing should follow some properties \cite{modeleditingsurvey}: reliability (predicting updated facts), locality, generality, and efficiency (efficient in runtime and memory).
The direct but inefficient editing is to fully finetune a model on new facts \cite{DBLP:journals/corr/abs-2012-00363}.
For locality, many works \citep{DBLP:conf/acl/DaiDHSCW22, ROME, MEMIT, DBLP:journals/corr/abs-2310-10322, alphaedit, anyedit} seek the model parameters strongly related to the facts and then edit these localized hidden states.
With high efficiency, \citet{MEND, malmen} achieve fast editing by training specific editor networks.
Also, lifelong model editing, like WISE \cite{wise}, is paid attention to for practical applications.
Recently, model editing has been applied to unlearn information from language models \cite{DBLP:journals/corr/abs-2309-17410, DBLP:journals/corr/abs-2309-11852, PCGU, washing}.
Inspired by them, we propose an efficient bias editing method, \textsc{BiasEdit}, to eliminate bias in language models while preserving the language modeling capabilities and generalizing gender reversal inputs and semantically related inputs.


\section{Conclusion}
We propose \textbf{\textsc{BiasEdit}}, an efficient model editing method to debias stereotyped language models by modifying a small portion of language models' parameters with small editor networks.
We design a debiasing loss $\mathcal{L}_d$ for debiasing and a retention loss $\mathcal{L}_r$ to maintain the language modeling abilities during editing.
Experiments illustrate that \textsc{BiasEdit} presents much better debiasing performance than classical debiasing methods and gives little to no harmful impact on language modeling and general capabilities.
Also, \ours\ is robust in gender reversal and semantic generality.
Meanwhile, we comprehensively investigate the effects of debiasing different components of language models.

\section*{Limitations}
\ours~ is only evaluated on sentence-level bias modeling examples with gold labels.
However, in the LLM era, we expect bias mitigation for text generation forms, such as QA and text continuation, which is more appropriate for current chat-based large language models.
Furthermore, biased datasets for text generation, like BBQ \cite{parrish-etal-2022-bbq}, with gold labels are extremely lacking.
Therefore, we hope that \ours~and other adapt model editing / unlearning methods can be adapted to mitigate bias for text generation, and such datasets will be constructed in the future.

\section*{Ethics Statement}
This work hopes to encourage more research for debiasing language models.
We use open-source pre-trained language models from \textit{HuggingFace} \cite{huggingface}.
All datasets and codes in the experiments are publicly available.
We ensure that no private information is in our research.
Furthermore, we recognize the potential societal impacts of our work that \ours\ can be immorally used to make language models more biased, which is harmful to society.
We advocate for the responsible use of our method in ways that benefit the whole society and minimize harm.

\section*{Acknowledgements}
This work was done during Xin Xu's internship mentored by Prof. Wei Xu.
Thanks for the code suggestions from Chenmien Tan, one author of MALMEN \cite{malmen}.

\bibliography{anthology,custom}

\begin{thebibliography}{98}
\providecommand{\natexlab}[1]{#1}

\bibitem[{Abdi and Williams(2010)}]{PCA}
H.~Abdi and L.~J. Williams. 2010.
\newblock \href {https://doi.org/10.1002/wics.101} {Principal component analysis}.
\newblock \emph{WIREs Computational Statistics}, 2:433--459.

\bibitem[{Barikeri et~al.(2021)Barikeri, Lauscher, Vuli{\'c}, and Glava{\v{s}}}]{barikeri-etal-2021-redditbias}
Soumya Barikeri, Anne Lauscher, Ivan Vuli{\'c}, and Goran Glava{\v{s}}. 2021.
\newblock \href {https://doi.org/10.18653/v1/2021.acl-long.151} {{R}eddit{B}ias: A real-world resource for bias evaluation and debiasing of conversational language models}.
\newblock In \emph{Proceedings of the 59th Annual Meeting of the Association for Computational Linguistics and the 11th International Joint Conference on Natural Language Processing (Volume 1: Long Papers)}, pages 1941--1955, Online. Association for Computational Linguistics.

\bibitem[{Bauer et~al.(2023)Bauer, Tischer, and Bansal}]{DBLP:conf/eacl/BauerTB23}
Lisa Bauer, Hanna Tischer, and Mohit Bansal. 2023.
\newblock \href {https://aclanthology.org/2023.eacl-main.271} {Social commonsense for explanation and cultural bias discovery}.
\newblock In \emph{Proceedings of the 17th Conference of the European Chapter of the Association for Computational Linguistics, {EACL} 2023, Dubrovnik, Croatia, May 2-6, 2023}, pages 3727--3742. Association for Computational Linguistics.

\bibitem[{Berglund et~al.(2023)Berglund, Tong, Kaufmann, Balesni, Stickland, Korbak, and Evans}]{reversalcurse}
Lukas Berglund, Meg Tong, Max Kaufmann, Mikita Balesni, Asa~Cooper Stickland, Tomasz Korbak, and Owain Evans. 2023.
\newblock \href {https://doi.org/10.48550/ARXIV.2309.12288} {The reversal curse: Llms trained on "a is b" fail to learn "b is a"}.
\newblock \emph{CoRR}, abs/2309.12288.

\bibitem[{Bird and Loper(2004)}]{NLTK}
Steven Bird and Edward Loper. 2004.
\newblock \href {https://aclanthology.org/P04-3031} {{NLTK}: The natural language toolkit}.
\newblock In \emph{Proceedings of the {ACL} Interactive Poster and Demonstration Sessions}, pages 214--217, Barcelona, Spain. Association for Computational Linguistics.

\bibitem[{Cao et~al.(2021)Cao, Aziz, and Titov}]{KE}
Nicola~De Cao, Wilker Aziz, and Ivan Titov. 2021.
\newblock \href {https://doi.org/10.18653/v1/2021.emnlp-main.522} {Editing factual knowledge in language models}.
\newblock In \emph{Proceedings of the 2021 Conference on Empirical Methods in Natural Language Processing, {EMNLP} 2021, Virtual Event / Punta Cana, Dominican Republic, 7-11 November, 2021}, pages 6491--6506. Association for Computational Linguistics.

\bibitem[{Cao et~al.(2022)Cao, Sotnikova, III, Rudinger, and Zou}]{DBLP:conf/naacl/CaoSDRZ22}
Yang~Trista Cao, Anna Sotnikova, Hal~Daum{\'{e}} III, Rachel Rudinger, and Linda Zou. 2022.
\newblock \href {https://doi.org/10.18653/V1/2022.NAACL-MAIN.92} {Theory-grounded measurement of {U.S.} social stereotypes in english language models}.
\newblock In \emph{Proceedings of the 2022 Conference of the North American Chapter of the Association for Computational Linguistics: Human Language Technologies, {NAACL} 2022, Seattle, WA, United States, July 10-15, 2022}, pages 1276--1295. Association for Computational Linguistics.

\bibitem[{Chen et~al.(2024)Chen, Li, Xiao, and Liu}]{DBLP:journals/corr/abs-2405-09341}
Ruizhe Chen, Yichen Li, Zikai Xiao, and Zuozhu Liu. 2024.
\newblock \href {https://doi.org/10.48550/ARXIV.2405.09341} {Large language model bias mitigation from the perspective of knowledge editing}.
\newblock \emph{CoRR}, abs/2405.09341.

\bibitem[{Cheng et~al.(2023{\natexlab{a}})Cheng, Durmus, and Jurafsky}]{DBLP:conf/acl/ChengDJ23}
Myra Cheng, Esin Durmus, and Dan Jurafsky. 2023{\natexlab{a}}.
\newblock \href {https://doi.org/10.18653/v1/2023.acl-long.84} {Marked personas: Using natural language prompts to measure stereotypes in language models}.
\newblock In \emph{Proceedings of the 61st Annual Meeting of the Association for Computational Linguistics (Volume 1: Long Papers), {ACL} 2023, Toronto, Canada, July 9-14, 2023}, pages 1504--1532. Association for Computational Linguistics.

\bibitem[{Cheng et~al.(2023{\natexlab{b}})Cheng, Zhang, Tian, Dai, Xiong, Guo, and Chen}]{siyuanhyper}
Siyuan Cheng, Ningyu Zhang, Bozhong Tian, Zelin Dai, Feiyu Xiong, Wei Guo, and Huajun Chen. 2023{\natexlab{b}}.
\newblock \href {https://doi.org/10.48550/ARXIV.2301.10405} {Editing language model-based knowledge graph embeddings}.
\newblock \emph{CoRR}, abs/2301.10405.

\bibitem[{Chintam et~al.(2023)Chintam, Beloch, Zuidema, Hanna, and van~der Wal}]{chintam-etal-2023-identifying}
Abhijith Chintam, Rahel Beloch, Willem Zuidema, Michael Hanna, and Oskar van~der Wal. 2023.
\newblock \href {https://doi.org/10.18653/v1/2023.blackboxnlp-1.29} {Identifying and adapting transformer-components responsible for gender bias in an {E}nglish language model}.
\newblock In \emph{Proceedings of the 6th BlackboxNLP Workshop: Analyzing and Interpreting Neural Networks for NLP}, pages 379--394, Singapore. Association for Computational Linguistics.

\bibitem[{Clark et~al.(2019)Clark, Lee, Chang, Kwiatkowski, Collins, and Toutanova}]{boolq}
Christopher Clark, Kenton Lee, Ming{-}Wei Chang, Tom Kwiatkowski, Michael Collins, and Kristina Toutanova. 2019.
\newblock \href {https://doi.org/10.18653/V1/N19-1300} {Boolq: Exploring the surprising difficulty of natural yes/no questions}.
\newblock In \emph{Proceedings of the 2019 Conference of the North American Chapter of the Association for Computational Linguistics: Human Language Technologies, {NAACL-HLT} 2019, Minneapolis, MN, USA, June 2-7, 2019, Volume 1 (Long and Short Papers)}, pages 2924--2936. Association for Computational Linguistics.

\bibitem[{Contributors(2023)}]{opencompass}
OpenCompass Contributors. 2023.
\newblock Opencompass: A universal evaluation platform for foundation models.
\newblock \url{https://github.com/open-compass/opencompass}.

\bibitem[{Dai et~al.(2022)Dai, Dong, Hao, Sui, Chang, and Wei}]{DBLP:conf/acl/DaiDHSCW22}
Damai Dai, Li~Dong, Yaru Hao, Zhifang Sui, Baobao Chang, and Furu Wei. 2022.
\newblock \href {https://doi.org/10.18653/V1/2022.ACL-LONG.581} {Knowledge neurons in pretrained transformers}.
\newblock In \emph{Proceedings of the 60th Annual Meeting of the Association for Computational Linguistics (Volume 1: Long Papers), {ACL} 2022, Dublin, Ireland, May 22-27, 2022}, pages 8493--8502. Association for Computational Linguistics.

\bibitem[{Das et~al.(2023)Das, Guha, and Semaan}]{das-etal-2023-toward}
Dipto Das, Shion Guha, and Bryan Semaan. 2023.
\newblock \href {https://doi.org/10.18653/v1/2023.c3nlp-1.8} {Toward cultural bias evaluation datasets: The case of {B}engali gender, religious, and national identity}.
\newblock In \emph{Proceedings of the First Workshop on Cross-Cultural Considerations in NLP (C3NLP)}, pages 68--83, Dubrovnik, Croatia. Association for Computational Linguistics.

\bibitem[{Devine(1989)}]{devine1989stereotypes}
Patricia~G Devine. 1989.
\newblock Stereotypes and prejudice: Their automatic and controlled components.
\newblock \emph{Journal of personality and social psychology}, 56(1):5.

\bibitem[{Faisal and Anastasopoulos(2022)}]{DBLP:journals/corr/abs-2212-10408}
Fahim Faisal and Antonios Anastasopoulos. 2022.
\newblock \href {https://doi.org/10.48550/ARXIV.2212.10408} {Geographic and geopolitical biases of language models}.
\newblock \emph{CoRR}, abs/2212.10408.

\bibitem[{Fang et~al.(2024)Fang, Jiang, Wang, Ma, Wang, He, and Chua}]{alphaedit}
Junfeng Fang, Houcheng Jiang, Kun Wang, Yunshan Ma, Xiang Wang, Xiangnan He, and Tat{-}Seng Chua. 2024.
\newblock \href {https://doi.org/10.48550/ARXIV.2410.02355} {Alphaedit: Null-space constrained knowledge editing for language models}.
\newblock \emph{CoRR}, abs/2410.02355.

\bibitem[{Gallegos et~al.(2023)Gallegos, Rossi, Barrow, Tanjim, Kim, Dernoncourt, Yu, Zhang, and Ahmed}]{DBLP:journals/corr/abs-2309-00770}
Isabel~O. Gallegos, Ryan~A. Rossi, Joe Barrow, Md.~Mehrab Tanjim, Sungchul Kim, Franck Dernoncourt, Tong Yu, Ruiyi Zhang, and Nesreen~K. Ahmed. 2023.
\newblock \href {https://doi.org/10.48550/ARXIV.2309.00770} {Bias and fairness in large language models: {A} survey}.
\newblock \emph{CoRR}, abs/2309.00770.

\bibitem[{Geva et~al.(2021)Geva, Schuster, Berant, and Levy}]{DBLP:conf/emnlp/GevaSBL21}
Mor Geva, Roei Schuster, Jonathan Berant, and Omer Levy. 2021.
\newblock \href {https://doi.org/10.18653/V1/2021.EMNLP-MAIN.446} {Transformer feed-forward layers are key-value memories}.
\newblock In \emph{Proceedings of the 2021 Conference on Empirical Methods in Natural Language Processing, {EMNLP} 2021, Virtual Event / Punta Cana, Dominican Republic, 7-11 November, 2021}, pages 5484--5495. Association for Computational Linguistics.

\bibitem[{Gira et~al.(2022)Gira, Zhang, and Lee}]{DBLP:conf/ltedi/GiraZL22}
Michael Gira, Ruisu Zhang, and Kangwook Lee. 2022.
\newblock \href {https://doi.org/10.18653/V1/2022.LTEDI-1.8} {Debiasing pre-trained language models via efficient fine-tuning}.
\newblock In \emph{Proceedings of the Second Workshop on Language Technology for Equality, Diversity and Inclusion, {LT-EDI} 2022, Dublin, Ireland, May 27, 2022}, pages 59--69. Association for Computational Linguistics.

\bibitem[{Goldfarb{-}Tarrant et~al.(2023)Goldfarb{-}Tarrant, Ungless, Balkir, and Blodgett}]{DBLP:conf/acl/Goldfarb-Tarrant23}
Seraphina Goldfarb{-}Tarrant, Eddie Ungless, Esma Balkir, and Su~Lin Blodgett. 2023.
\newblock \href {https://doi.org/10.18653/V1/2023.FINDINGS-ACL.139} {This prompt is measuring {\textbackslash}textlessmask{\textbackslash}textgreater: evaluating bias evaluation in language models}.
\newblock In \emph{Findings of the Association for Computational Linguistics: {ACL} 2023, Toronto, Canada, July 9-14, 2023}, pages 2209--2225. Association for Computational Linguistics.

\bibitem[{Gu et~al.(2024)Gu, Xu, Ma, Lu, Ling, Chang, and Peng}]{DBLP:journals/corr/abs-2401-04700}
Jia{-}Chen Gu, Hao{-}Xiang Xu, Jun{-}Yu Ma, Pan Lu, Zhen{-}Hua Ling, Kai{-}Wei Chang, and Nanyun Peng. 2024.
\newblock \href {https://doi.org/10.48550/ARXIV.2401.04700} {Model editing can hurt general abilities of large language models}.
\newblock \emph{CoRR}, abs/2401.04700.

\bibitem[{Guo et~al.(2022)Guo, Yang, and Abbasi}]{auto-debias}
Yue Guo, Yi~Yang, and Ahmed Abbasi. 2022.
\newblock \href {https://doi.org/10.18653/V1/2022.ACL-LONG.72} {Auto-debias: Debiasing masked language models with automated biased prompts}.
\newblock In \emph{Proceedings of the 60th Annual Meeting of the Association for Computational Linguistics (Volume 1: Long Papers), {ACL} 2022, Dublin, Ireland, May 22-27, 2022}, pages 1012--1023. Association for Computational Linguistics.

\bibitem[{Gupta et~al.(2024)Gupta, Rao, and Anumanchipalli}]{DBLP:journals/corr/abs-2401-07453}
Akshat Gupta, Anurag Rao, and Gopala Anumanchipalli. 2024.
\newblock \href {https://doi.org/10.48550/ARXIV.2401.07453} {Model editing at scale leads to gradual and catastrophic forgetting}.
\newblock \emph{CoRR}, abs/2401.07453.

\bibitem[{Gupta et~al.(2023)Gupta, Mondal, Sheshadri, Zhao, Li, Wiegreffe, and Tandon}]{DBLP:conf/emnlp/GuptaMS00WT23}
Anshita Gupta, Debanjan Mondal, Akshay~Krishna Sheshadri, Wenlong Zhao, Xiang Li, Sarah Wiegreffe, and Niket Tandon. 2023.
\newblock \href {https://aclanthology.org/2023.emnlp-main.511} {Editing common sense in transformers}.
\newblock In \emph{Proceedings of the 2023 Conference on Empirical Methods in Natural Language Processing, {EMNLP} 2023, Singapore, December 6-10, 2023}, pages 8214--8232. Association for Computational Linguistics.

\bibitem[{Halevy et~al.(2021)Halevy, Harris, Bruckman, Yang, and Howard}]{DBLP:conf/eaamo/HalevyHBYH21}
Matan Halevy, Camille Harris, Amy~S. Bruckman, Diyi Yang, and Ayanna~M. Howard. 2021.
\newblock \href {https://doi.org/10.1145/3465416.3483299} {Mitigating racial biases in toxic language detection with an equity-based ensemble framework}.
\newblock In \emph{{EAAMO} 2021: {ACM} Conference on Equity and Access in Algorithms, Mechanisms, and Optimization, Virtual Event, USA, October 5 - 9, 2021}, pages 7:1--7:11. {ACM}.

\bibitem[{Hase et~al.(2023)Hase, Diab, Celikyilmaz, Li, Kozareva, Stoyanov, Bansal, and Iyer}]{DBLP:conf/eacl/HaseDCLKSBI23}
Peter Hase, Mona~T. Diab, Asli Celikyilmaz, Xian Li, Zornitsa Kozareva, Veselin Stoyanov, Mohit Bansal, and Srinivasan Iyer. 2023.
\newblock \href {https://doi.org/10.18653/V1/2023.EACL-MAIN.199} {Methods for measuring, updating, and visualizing factual beliefs in language models}.
\newblock In \emph{Proceedings of the 17th Conference of the European Chapter of the Association for Computational Linguistics, {EACL} 2023, Dubrovnik, Croatia, May 2-6, 2023}, pages 2706--2723. Association for Computational Linguistics.

\bibitem[{Hovy and Prabhumoye(2021)}]{DBLP:journals/llc/HovyP21}
Dirk Hovy and Shrimai Prabhumoye. 2021.
\newblock \href {https://doi.org/10.1111/LNC3.12432} {Five sources of bias in natural language processing}.
\newblock \emph{Lang. Linguistics Compass}, 15(8).

\bibitem[{Ishibashi and Shimodaira(2023)}]{DBLP:journals/corr/abs-2309-11852}
Yoichi Ishibashi and Hidetoshi Shimodaira. 2023.
\newblock \href {https://doi.org/10.48550/ARXIV.2309.11852} {Knowledge sanitization of large language models}.
\newblock \emph{CoRR}, abs/2309.11852.

\bibitem[{Iskander et~al.(2023)Iskander, Radinsky, and Belinkov}]{DBLP:conf/acl/IskanderRB23}
Shadi Iskander, Kira Radinsky, and Yonatan Belinkov. 2023.
\newblock \href {https://doi.org/10.18653/V1/2023.FINDINGS-ACL.369} {Shielded representations: Protecting sensitive attributes through iterative gradient-based projection}.
\newblock In \emph{Findings of the Association for Computational Linguistics: {ACL} 2023, Toronto, Canada, July 9-14, 2023}, pages 5961--5977. Association for Computational Linguistics.

\bibitem[{Jiang et~al.(2023)Jiang, Sablayrolles, Mensch, Bamford, Chaplot, de~Las~Casas, Bressand, Lengyel, Lample, Saulnier, Lavaud, Lachaux, Stock, Scao, Lavril, Wang, Lacroix, and Sayed}]{mistral}
Albert~Q. Jiang, Alexandre Sablayrolles, Arthur Mensch, Chris Bamford, Devendra~Singh Chaplot, Diego de~Las~Casas, Florian Bressand, Gianna Lengyel, Guillaume Lample, Lucile Saulnier, L{\'{e}}lio~Renard Lavaud, Marie{-}Anne Lachaux, Pierre Stock, Teven~Le Scao, Thibaut Lavril, Thomas Wang, Timoth{\'{e}}e Lacroix, and William~El Sayed. 2023.
\newblock \href {https://doi.org/10.48550/ARXIV.2310.06825} {Mistral 7b}.
\newblock \emph{CoRR}, abs/2310.06825.

\bibitem[{Jiang et~al.(2025)Jiang, Fang, Zhang, Ma, Wan, Wang, He, and Chua}]{anyedit}
Houcheng Jiang, Junfeng Fang, Ningyu Zhang, Guojun Ma, Mingyang Wan, Xiang Wang, Xiangnan He, and Tat-seng Chua. 2025.
\newblock Anyedit: Edit any knowledge encoded in language models.
\newblock \emph{arXiv preprint arXiv:2502.05628}.

\bibitem[{Jiang et~al.(2020)Jiang, Xu, Araki, and Neubig}]{DBLP:journals/tacl/JiangXAN20}
Zhengbao Jiang, Frank~F. Xu, Jun Araki, and Graham Neubig. 2020.
\newblock \href {https://doi.org/10.1162/TACL\_A\_00324} {How can we know what language models know}.
\newblock \emph{Trans. Assoc. Comput. Linguistics}, 8:423--438.

\bibitem[{Joniak and Aizawa(2022)}]{DBLP:journals/corr/abs-2207-02463}
Przemyslaw~K. Joniak and Akiko Aizawa. 2022.
\newblock \href {https://doi.org/10.48550/ARXIV.2207.02463} {Gender biases and where to find them: Exploring gender bias in pre-trained transformer-based language models using movement pruning}.
\newblock \emph{CoRR}, abs/2207.02463.

\bibitem[{Kumar et~al.(2023)Kumar, Lesota, Zerveas, Cohen, Eickhoff, Schedl, and Rekabsaz}]{kumar-etal-2023-parameter}
Deepak Kumar, Oleg Lesota, George Zerveas, Daniel Cohen, Carsten Eickhoff, Markus Schedl, and Navid Rekabsaz. 2023.
\newblock \href {https://doi.org/10.18653/v1/2023.eacl-main.201} {Parameter-efficient modularised bias mitigation via {A}dapter{F}usion}.
\newblock In \emph{Proceedings of the 17th Conference of the European Chapter of the Association for Computational Linguistics}, pages 2738--2751, Dubrovnik, Croatia. Association for Computational Linguistics.

\bibitem[{Li et~al.(2024)Li, Chen, and Wang}]{DBLP:journals/corr/abs-2401-02976}
Jiahang Li, Taoyu Chen, and Yuanli Wang. 2024.
\newblock \href {https://doi.org/10.48550/ARXIV.2401.02976} {Trace and edit relation associations in {GPT}}.
\newblock \emph{CoRR}, abs/2401.02976.

\bibitem[{Li et~al.(2022{\natexlab{a}})Li, Li, Shang, Dong, Sun, Liu, Ji, Jiang, and Liu}]{DBLP:conf/acl/LiLSDSLJJL22}
Shaobo Li, Xiaoguang Li, Lifeng Shang, Zhenhua Dong, Chengjie Sun, Bingquan Liu, Zhenzhou Ji, Xin Jiang, and Qun Liu. 2022{\natexlab{a}}.
\newblock \href {https://doi.org/10.18653/V1/2022.FINDINGS-ACL.136} {How pre-trained language models capture factual knowledge? {A} causal-inspired analysis}.
\newblock In \emph{Findings of the Association for Computational Linguistics: {ACL} 2022, Dublin, Ireland, May 22-27, 2022}, pages 1720--1732. Association for Computational Linguistics.

\bibitem[{Li et~al.(2022{\natexlab{b}})Li, Zhang, Yang, Lin, Ragni, Wang, and Fu}]{DBLP:conf/ijcnlp/LiZYLRWF22}
Yizhi Li, Ge~Zhang, Bohao Yang, Chenghua Lin, Anton Ragni, Shi Wang, and Jie Fu. 2022{\natexlab{b}}.
\newblock \href {https://aclanthology.org/2022.findings-aacl.32} {{HERB:} measuring hierarchical regional bias in pre-trained language models}.
\newblock In \emph{Findings of the Association for Computational Linguistics: {AACL-IJCNLP} 2022, Online only, November 20-23, 2022}, pages 334--346. Association for Computational Linguistics.

\bibitem[{Liang et~al.(2020)Liang, Li, Zheng, Lim, Salakhutdinov, and Morency}]{sentencedebias}
Paul~Pu Liang, Irene~Mengze Li, Emily Zheng, Yao~Chong Lim, Ruslan Salakhutdinov, and Louis{-}Philippe Morency. 2020.
\newblock \href {https://doi.org/10.18653/v1/2020.acl-main.488} {Towards debiasing sentence representations}.
\newblock In \emph{Proceedings of the 58th Annual Meeting of the Association for Computational Linguistics, {ACL} 2020, Online, July 5-10, 2020}, pages 5502--5515. Association for Computational Linguistics.

\bibitem[{Liang et~al.(2021)Liang, Wu, Morency, and Salakhutdinov}]{DBLP:conf/icml/LiangWMS21}
Paul~Pu Liang, Chiyu Wu, Louis{-}Philippe Morency, and Ruslan Salakhutdinov. 2021.
\newblock \href {http://proceedings.mlr.press/v139/liang21a.html} {Towards understanding and mitigating social biases in language models}.
\newblock In \emph{Proceedings of the 38th International Conference on Machine Learning, {ICML} 2021, 18-24 July 2021, Virtual Event}, volume 139 of \emph{Proceedings of Machine Learning Research}, pages 6565--6576. {PMLR}.

\bibitem[{Limisiewicz and Marecek(2022)}]{DBLP:journals/corr/abs-2206-10744}
Tomasz Limisiewicz and David Marecek. 2022.
\newblock \href {https://doi.org/10.48550/ARXIV.2206.10744} {Don't forget about pronouns: Removing gender bias in language models without losing factual gender information}.
\newblock \emph{CoRR}, abs/2206.10744.

\bibitem[{Limisiewicz et~al.(2024)Limisiewicz, Marecek, and Musil}]{DAMA}
Tomasz Limisiewicz, David Marecek, and Tom{\'{a}}s Musil. 2024.
\newblock \href {https://openreview.net/forum?id=XIZEFyVGC9} {Debiasing algorithm through model adaptation}.
\newblock In \emph{The Twelfth International Conference on Learning Representations, {ICLR} 2024, Vienna, Austria, May 7-11, 2024}. OpenReview.net.

\bibitem[{Lin et~al.(2025)Lin, Basu, Beigi, Manjunatha, Rossi, Wang, Zhou, Balasubramanian, Zarei, Rezaei et~al.}]{lin2025survey}
Zihao Lin, Samyadeep Basu, Mohammad Beigi, Varun Manjunatha, Ryan~A Rossi, Zichao Wang, Yufan Zhou, Sriram Balasubramanian, Arman Zarei, Keivan Rezaei, et~al. 2025.
\newblock A survey on mechanistic interpretability for multi-modal foundation models.
\newblock \emph{arXiv preprint arXiv:2502.17516}.

\bibitem[{Liu et~al.(2023)Liu, Yao, Ton, Zhang, Guo, Cheng, Klochkov, Taufiq, and Li}]{TrustLLM}
Yang Liu, Yuanshun Yao, Jean{-}Francois Ton, Xiaoying Zhang, Ruocheng Guo, Hao Cheng, Yegor Klochkov, Muhammad~Faaiz Taufiq, and Hang Li. 2023.
\newblock \href {https://doi.org/10.48550/ARXIV.2308.05374} {Trustworthy llms: a survey and guideline for evaluating large language models' alignment}.
\newblock \emph{CoRR}, abs/2308.05374.

\bibitem[{Lu et~al.(2020)Lu, Mardziel, Wu, Amancharla, and Datta}]{DBLP:conf/birthday/LuMWAD20}
Kaiji Lu, Piotr Mardziel, Fangjing Wu, Preetam Amancharla, and Anupam Datta. 2020.
\newblock \href {https://doi.org/10.1007/978-3-030-62077-6\_14} {Gender bias in neural natural language processing}.
\newblock In \emph{Logic, Language, and Security - Essays Dedicated to Andre Scedrov on the Occasion of His 65th Birthday}, volume 12300 of \emph{Lecture Notes in Computer Science}, pages 189--202. Springer.

\bibitem[{Ma et~al.(2023{\natexlab{a}})Ma, Gu, Ling, Liu, and Liu}]{DBLP:journals/corr/abs-2310-10322}
Jun{-}Yu Ma, Jia{-}Chen Gu, Zhen{-}Hua Ling, Quan Liu, and Cong Liu. 2023{\natexlab{a}}.
\newblock \href {https://doi.org/10.48550/ARXIV.2310.10322} {Untying the reversal curse via bidirectional language model editing}.
\newblock \emph{CoRR}, abs/2310.10322.

\bibitem[{Ma et~al.(2023{\natexlab{b}})Ma, Scheible, Wang, Veeramachaneni, Chowdhary, Sun, Koulogeorge, Wang, Yang, and Vosoughi}]{DBLP:conf/emnlp/MaSWVCSKWYV23}
Weicheng Ma, Henry Scheible, Brian Wang, Goutham Veeramachaneni, Pratim Chowdhary, Alan Sun, Andrew Koulogeorge, Lili Wang, Diyi Yang, and Soroush Vosoughi. 2023{\natexlab{b}}.
\newblock \href {https://aclanthology.org/2023.emnlp-main.697} {Deciphering stereotypes in pre-trained language models}.
\newblock In \emph{Proceedings of the 2023 Conference on Empirical Methods in Natural Language Processing, {EMNLP} 2023, Singapore, December 6-10, 2023}, pages 11328--11345. Association for Computational Linguistics.

\bibitem[{Manzini et~al.(2019)Manzini, Yao~Chong, Black, and Tsvetkov}]{manzini-etal-2019-black}
Thomas Manzini, Lim Yao~Chong, Alan~W Black, and Yulia Tsvetkov. 2019.
\newblock \href {https://doi.org/10.18653/v1/N19-1062} {{B}lack is to criminal as {C}aucasian is to police: Detecting and removing multiclass bias in word embeddings}.
\newblock In \emph{Proceedings of the 2019 Conference of the North {A}merican Chapter of the Association for Computational Linguistics: Human Language Technologies, Volume 1 (Long and Short Papers)}, pages 615--621, Minneapolis, Minnesota. Association for Computational Linguistics.

\bibitem[{Mattern et~al.(2022)Mattern, Jin, Sachan, Mihalcea, and Sch{\"{o}}lkopf}]{DBLP:journals/corr/abs-2212-10678}
Justus Mattern, Zhijing Jin, Mrinmaya Sachan, Rada Mihalcea, and Bernhard Sch{\"{o}}lkopf. 2022.
\newblock \href {https://doi.org/10.48550/ARXIV.2212.10678} {Understanding stereotypes in language models: Towards robust measurement and zero-shot debiasing}.
\newblock \emph{CoRR}, abs/2212.10678.

\bibitem[{Meade et~al.(2022)Meade, Poole{-}Dayan, and Reddy}]{bias-bench}
Nicholas Meade, Elinor Poole{-}Dayan, and Siva Reddy. 2022.
\newblock \href {https://doi.org/10.18653/v1/2022.acl-long.132} {An empirical survey of the effectiveness of debiasing techniques for pre-trained language models}.
\newblock In \emph{Proceedings of the 60th Annual Meeting of the Association for Computational Linguistics (Volume 1: Long Papers), {ACL} 2022, Dublin, Ireland, May 22-27, 2022}, pages 1878--1898. Association for Computational Linguistics.

\bibitem[{Meng et~al.(2022)Meng, Bau, Andonian, and Belinkov}]{ROME}
Kevin Meng, David Bau, Alex Andonian, and Yonatan Belinkov. 2022.
\newblock \href {http://papers.nips.cc/paper\_files/paper/2022/hash/6f1d43d5a82a37e89b0665b33bf3a182-Abstract-Conference.html} {Locating and editing factual associations in {GPT}}.
\newblock In \emph{NeurIPS}.

\bibitem[{Meng et~al.(2023)Meng, Sharma, Andonian, Belinkov, and Bau}]{MEMIT}
Kevin Meng, Arnab~Sen Sharma, Alex~J. Andonian, Yonatan Belinkov, and David Bau. 2023.
\newblock \href {https://openreview.net/pdf?id=MkbcAHIYgyS} {Mass-editing memory in a transformer}.
\newblock In \emph{The Eleventh International Conference on Learning Representations, {ICLR} 2023, Kigali, Rwanda, May 1-5, 2023}. OpenReview.net.

\bibitem[{Merullo et~al.(2023)Merullo, Eickhoff, and Pavlick}]{DBLP:journals/corr/abs-2305-16130}
Jack Merullo, Carsten Eickhoff, and Ellie Pavlick. 2023.
\newblock \href {https://doi.org/10.48550/ARXIV.2305.16130} {Language models implement simple word2vec-style vector arithmetic}.
\newblock \emph{CoRR}, abs/2305.16130.

\bibitem[{Mesnard et~al.(2024)Mesnard, Hardin, Dadashi, Bhupatiraju, Pathak, Sifre, Rivi{\`{e}}re, Kale, Love, Tafti, Hussenot, Chowdhery, Roberts, Barua, Botev, Castro{-}Ros, Slone, H{\'{e}}liou, Tacchetti, Bulanova, Paterson, Tsai, Shahriari, Lan, Choquette{-}Choo, Crepy, Cer, Ippolito, Reid, Buchatskaya, Ni, Noland, Yan, Tucker, Muraru, Rozhdestvenskiy, Michalewski, Tenney, Grishchenko, Austin, Keeling, Labanowski, Lespiau, Stanway, Brennan, Chen, Ferret, Chiu, and et~al.}]{gemma}
Thomas Mesnard, Cassidy Hardin, Robert Dadashi, Surya Bhupatiraju, Shreya Pathak, Laurent Sifre, Morgane Rivi{\`{e}}re, Mihir~Sanjay Kale, Juliette Love, Pouya Tafti, L{\'{e}}onard Hussenot, Aakanksha Chowdhery, Adam Roberts, Aditya Barua, Alex Botev, Alex Castro{-}Ros, Ambrose Slone, Am{\'{e}}lie H{\'{e}}liou, Andrea Tacchetti, Anna Bulanova, Antonia Paterson, Beth Tsai, Bobak Shahriari, Charline~Le Lan, Christopher~A. Choquette{-}Choo, Cl{\'{e}}ment Crepy, Daniel Cer, Daphne Ippolito, David Reid, Elena Buchatskaya, Eric Ni, Eric Noland, Geng Yan, George Tucker, George{-}Cristian Muraru, Grigory Rozhdestvenskiy, Henryk Michalewski, Ian Tenney, Ivan Grishchenko, Jacob Austin, James Keeling, Jane Labanowski, Jean{-}Baptiste Lespiau, Jeff Stanway, Jenny Brennan, Jeremy Chen, Johan Ferret, Justin Chiu, and et~al. 2024.
\newblock \href {https://doi.org/10.48550/ARXIV.2403.08295} {Gemma: Open models based on gemini research and technology}.
\newblock \emph{CoRR}, abs/2403.08295.

\bibitem[{Meta(2024)}]{llama3}
Meta. 2024.
\newblock Introducing meta llama 3: The most capable openly available llm to date.
\newblock \url{https://ai.meta.com/blog/meta-llama-3/}.

\bibitem[{Mihaylov et~al.(2018)Mihaylov, Clark, Khot, and Sabharwal}]{openbookqa}
Todor Mihaylov, Peter Clark, Tushar Khot, and Ashish Sabharwal. 2018.
\newblock \href {https://doi.org/10.18653/V1/D18-1260} {Can a suit of armor conduct electricity? {A} new dataset for open book question answering}.
\newblock In \emph{Proceedings of the 2018 Conference on Empirical Methods in Natural Language Processing, Brussels, Belgium, October 31 - November 4, 2018}, pages 2381--2391. Association for Computational Linguistics.

\bibitem[{Miller(1995)}]{DBLP:journals/cacm/Miller95}
George~A. Miller. 1995.
\newblock \href {https://doi.org/10.1145/219717.219748} {Wordnet: {A} lexical database for english}.
\newblock \emph{Commun. {ACM}}, 38(11):39--41.

\bibitem[{Mitchell et~al.(2022)Mitchell, Lin, Bosselut, Finn, and Manning}]{MEND}
Eric Mitchell, Charles Lin, Antoine Bosselut, Chelsea Finn, and Christopher~D. Manning. 2022.
\newblock \href {https://openreview.net/forum?id=0DcZxeWfOPt} {Fast model editing at scale}.
\newblock In \emph{The Tenth International Conference on Learning Representations, {ICLR} 2022, Virtual Event, April 25-29, 2022}. OpenReview.net.

\bibitem[{Nadeem et~al.(2021)Nadeem, Bethke, and Reddy}]{stereoset}
Moin Nadeem, Anna Bethke, and Siva Reddy. 2021.
\newblock \href {https://doi.org/10.18653/v1/2021.acl-long.416} {Stereoset: Measuring stereotypical bias in pretrained language models}.
\newblock In \emph{Proceedings of the 59th Annual Meeting of the Association for Computational Linguistics and the 11th International Joint Conference on Natural Language Processing, {ACL/IJCNLP} 2021, (Volume 1: Long Papers), Virtual Event, August 1-6, 2021}, pages 5356--5371. Association for Computational Linguistics.

\bibitem[{Nangia et~al.(2020)Nangia, Vania, Bhalerao, and Bowman}]{crows-pairs}
Nikita Nangia, Clara Vania, Rasika Bhalerao, and Samuel~R. Bowman. 2020.
\newblock \href {https://doi.org/10.18653/v1/2020.emnlp-main.154} {Crows-pairs: {A} challenge dataset for measuring social biases in masked language models}.
\newblock In \emph{Proceedings of the 2020 Conference on Empirical Methods in Natural Language Processing, {EMNLP} 2020, Online, November 16-20, 2020}, pages 1953--1967. Association for Computational Linguistics.

\bibitem[{Ni et~al.(2023)Ni, Chen, Li, Hu, Xu, and Yang}]{DBLP:journals/corr/abs-2311-08011}
Shiwen Ni, Dingwei Chen, Chengming Li, Xiping Hu, Ruifeng Xu, and Min Yang. 2023.
\newblock \href {https://doi.org/10.48550/ARXIV.2311.08011} {Forgetting before learning: Utilizing parametric arithmetic for knowledge updating in large language models}.
\newblock \emph{CoRR}, abs/2311.08011.

\bibitem[{Omrani et~al.(2023)Omrani, Ziabari, Yu, Golazizian, Kennedy, Atari, Ji, and Dehghani}]{DBLP:conf/acl/OmraniZYGKAJD23}
Ali Omrani, Alireza~Salkhordeh Ziabari, Charles Yu, Preni Golazizian, Brendan Kennedy, Mohammad Atari, Heng Ji, and Morteza Dehghani. 2023.
\newblock \href {https://doi.org/10.18653/V1/2023.ACL-LONG.227} {Social-group-agnostic bias mitigation via the stereotype content model}.
\newblock In \emph{Proceedings of the 61st Annual Meeting of the Association for Computational Linguistics (Volume 1: Long Papers), {ACL} 2023, Toronto, Canada, July 9-14, 2023}, pages 4123--4139. Association for Computational Linguistics.

\bibitem[{Parrish et~al.(2022)Parrish, Chen, Nangia, Padmakumar, Phang, Thompson, Htut, and Bowman}]{parrish-etal-2022-bbq}
Alicia Parrish, Angelica Chen, Nikita Nangia, Vishakh Padmakumar, Jason Phang, Jana Thompson, Phu~Mon Htut, and Samuel Bowman. 2022.
\newblock \href {https://doi.org/10.18653/v1/2022.findings-acl.165} {{BBQ}: A hand-built bias benchmark for question answering}.
\newblock In \emph{Findings of the Association for Computational Linguistics: ACL 2022}, pages 2086--2105, Dublin, Ireland. Association for Computational Linguistics.

\bibitem[{Patil et~al.(2023)Patil, Hase, and Bansal}]{DBLP:journals/corr/abs-2309-17410}
Vaidehi Patil, Peter Hase, and Mohit Bansal. 2023.
\newblock \href {https://doi.org/10.48550/ARXIV.2309.17410} {Can sensitive information be deleted from llms? objectives for defending against extraction attacks}.
\newblock \emph{CoRR}, abs/2309.17410.

\bibitem[{Petroni et~al.(2019)Petroni, Rockt{\"{a}}schel, Riedel, Lewis, Bakhtin, Wu, and Miller}]{LAMA}
Fabio Petroni, Tim Rockt{\"{a}}schel, Sebastian Riedel, Patrick S.~H. Lewis, Anton Bakhtin, Yuxiang Wu, and Alexander~H. Miller. 2019.
\newblock \href {https://doi.org/10.18653/V1/D19-1250} {Language models as knowledge bases?}
\newblock In \emph{Proceedings of the 2019 Conference on Empirical Methods in Natural Language Processing and the 9th International Joint Conference on Natural Language Processing, {EMNLP-IJCNLP} 2019, Hong Kong, China, November 3-7, 2019}, pages 2463--2473. Association for Computational Linguistics.

\bibitem[{Radford et~al.(2019)Radford, Wu, Child, Luan, Amodei, and Sutskever}]{gpt2}
Alec Radford, Jeff Wu, Rewon Child, David Luan, Dario Amodei, and Ilya Sutskever. 2019.
\newblock Language models are unsupervised multitask learners.
\newblock \emph{OpenAI}.

\bibitem[{Ravfogel et~al.(2020)Ravfogel, Elazar, Gonen, Twiton, and Goldberg}]{INLP}
Shauli Ravfogel, Yanai Elazar, Hila Gonen, Michael Twiton, and Yoav Goldberg. 2020.
\newblock \href {https://doi.org/10.18653/v1/2020.acl-main.647} {Null it out: Guarding protected attributes by iterative nullspace projection}.
\newblock In \emph{Proceedings of the 58th Annual Meeting of the Association for Computational Linguistics}, pages 7237--7256, Online. Association for Computational Linguistics.

\bibitem[{Roemmele et~al.(2011)Roemmele, Bejan, and Gordon}]{copa}
Melissa Roemmele, Cosmin~Adrian Bejan, and Andrew~S. Gordon. 2011.
\newblock \href {http://www.aaai.org/ocs/index.php/SSS/SSS11/paper/view/2418} {Choice of plausible alternatives: An evaluation of commonsense causal reasoning}.
\newblock In \emph{Logical Formalizations of Commonsense Reasoning, Papers from the 2011 {AAAI} Spring Symposium, Technical Report SS-11-06, Stanford, California, USA, March 21-23, 2011}. {AAAI}.

\bibitem[{Schick et~al.(2021)Schick, Udupa, and Sch{\"{u}}tze}]{self-debias}
Timo Schick, Sahana Udupa, and Hinrich Sch{\"{u}}tze. 2021.
\newblock \href {https://doi.org/10.1162/tacl\_a\_00434} {Self-diagnosis and self-debiasing: {A} proposal for reducing corpus-based bias in {NLP}}.
\newblock \emph{Trans. Assoc. Comput. Linguistics}, 9:1408--1424.

\bibitem[{Sharkey et~al.(2025)Sharkey, Chughtai, Batson, Lindsey, Wu, Bushnaq, Goldowsky{-}Dill, Heimersheim, Ortega, Bloom, Biderman, Garriga{-}Alonso, Conmy, Nanda, Rumbelow, Wattenberg, Schoots, Miller, Michaud, Casper, Tegmark, Saunders, Bau, Todd, Geiger, Geva, Hoogland, Murfet, and McGrath}]{DBLP:journals/corr/abs-2501-16496}
Lee Sharkey, Bilal Chughtai, Joshua Batson, Jack Lindsey, Jeff Wu, Lucius Bushnaq, Nicholas Goldowsky{-}Dill, Stefan Heimersheim, Alejandro Ortega, Joseph~Isaac Bloom, Stella Biderman, Adri{\`{a}} Garriga{-}Alonso, Arthur Conmy, Neel Nanda, Jessica Rumbelow, Martin Wattenberg, Nandi Schoots, Joseph Miller, Eric~J. Michaud, Stephen Casper, Max Tegmark, William Saunders, David Bau, Eric Todd, Atticus Geiger, Mor Geva, Jesse Hoogland, Daniel Murfet, and Tom McGrath. 2025.
\newblock \href {https://doi.org/10.48550/ARXIV.2501.16496} {Open problems in mechanistic interpretability}.
\newblock \emph{CoRR}, abs/2501.16496.

\bibitem[{Sheng et~al.(2020)Sheng, Chang, Natarajan, and Peng}]{DBLP:conf/emnlp/ShengCNP20}
Emily Sheng, Kai{-}Wei Chang, Prem Natarajan, and Nanyun Peng. 2020.
\newblock \href {https://doi.org/10.18653/V1/2020.FINDINGS-EMNLP.291} {Towards controllable biases in language generation}.
\newblock In \emph{Findings of the Association for Computational Linguistics: {EMNLP} 2020, Online Event, 16-20 November 2020}, volume {EMNLP} 2020 of \emph{Findings of {ACL}}, pages 3239--3254. Association for Computational Linguistics.

\bibitem[{Shin et~al.(2020)Shin, Razeghi, IV, Wallace, and Singh}]{DBLP:conf/emnlp/ShinRLWS20}
Taylor Shin, Yasaman Razeghi, Robert L.~Logan IV, Eric Wallace, and Sameer Singh. 2020.
\newblock \href {https://doi.org/10.18653/V1/2020.EMNLP-MAIN.346} {Autoprompt: Eliciting knowledge from language models with automatically generated prompts}.
\newblock In \emph{Proceedings of the 2020 Conference on Empirical Methods in Natural Language Processing, {EMNLP} 2020, Online, November 16-20, 2020}, pages 4222--4235. Association for Computational Linguistics.

\bibitem[{Sinitsin et~al.(2020)Sinitsin, Plokhotnyuk, Pyrkin, Popov, and Babenko}]{ENN}
Anton Sinitsin, Vsevolod Plokhotnyuk, Dmitry~V. Pyrkin, Sergei Popov, and Artem Babenko. 2020.
\newblock \href {https://openreview.net/forum?id=HJedXaEtvS} {Editable neural networks}.
\newblock In \emph{8th International Conference on Learning Representations, {ICLR} 2020, Addis Ababa, Ethiopia, April 26-30, 2020}. OpenReview.net.

\bibitem[{Smith et~al.(2022)Smith, Hall, Kambadur, Presani, and Williams}]{DBLP:conf/emnlp/SmithHKPW22}
Eric~Michael Smith, Melissa Hall, Melanie Kambadur, Eleonora Presani, and Adina Williams. 2022.
\newblock \href {https://doi.org/10.18653/v1/2022.emnlp-main.625} {"i'm sorry to hear that": Finding new biases in language models with a holistic descriptor dataset}.
\newblock In \emph{Proceedings of the 2022 Conference on Empirical Methods in Natural Language Processing, {EMNLP} 2022, Abu Dhabi, United Arab Emirates, December 7-11, 2022}, pages 9180--9211. Association for Computational Linguistics.

\bibitem[{Sun et~al.(2019)Sun, Gaut, Tang, Huang, ElSherief, Zhao, Mirza, Belding, Chang, and Wang}]{DBLP:conf/acl/SunGTHEZMBCW19}
Tony Sun, Andrew Gaut, Shirlyn Tang, Yuxin Huang, Mai ElSherief, Jieyu Zhao, Diba Mirza, Elizabeth~M. Belding, Kai{-}Wei Chang, and William~Yang Wang. 2019.
\newblock \href {https://doi.org/10.18653/v1/p19-1159} {Mitigating gender bias in natural language processing: Literature review}.
\newblock In \emph{Proceedings of the 57th Conference of the Association for Computational Linguistics, {ACL} 2019, Florence, Italy, July 28- August 2, 2019, Volume 1: Long Papers}, pages 1630--1640. Association for Computational Linguistics.

\bibitem[{Tan et~al.(2023)Tan, Zhang, and Fu}]{malmen}
Chenmien Tan, Ge~Zhang, and Jie Fu. 2023.
\newblock \href {https://doi.org/10.48550/ARXIV.2311.04661} {Massive editing for large language models via meta learning}.
\newblock \emph{CoRR}, abs/2311.04661.

\bibitem[{Templeton et~al.(2024)Templeton, Conerly, Marcus, Lindsey, Bricken, Chen, Pearce, Citro, Ameisen, Jones, Cunningham, Turner, McDougall, MacDiarmid, Tamkin, Durmus, Hume, Mosconi, Freeman, Sumers, Rees, Batson, Jermyn, Carter, Olah, and Henighan}]{claude}
Adly Templeton, Tom Conerly, Jonathan Marcus, Jack Lindsey, Trenton Bricken, Brian Chen, Adam Pearce, Craig Citro, Emmanuel Ameisen, Andy Jones, Hoagy Cunningham, Nicholas~L Turner, Callum McDougall, Monte MacDiarmid, Alex Tamkin, Esin Durmus, Tristan Hume, Francesco Mosconi, C.~Daniel Freeman, Theodore~R. Sumers, Edward Rees, Joshua Batson, Adam Jermyn, Shan Carter, Chris Olah, and Tom Henighan. 2024.
\newblock Scaling monosemanticity: Extracting interpretable features from claude 3 sonnet.
\newblock \url{https://transformer-circuits.pub/2024/scaling-monosemanticity/index.html}.

\bibitem[{Vashishtha et~al.(2023)Vashishtha, Ahuja, and Sitaram}]{DBLP:conf/acl/VashishthaAS23}
Aniket Vashishtha, Kabir Ahuja, and Sunayana Sitaram. 2023.
\newblock \href {https://doi.org/10.18653/V1/2023.FINDINGS-ACL.21} {On evaluating and mitigating gender biases in multilingual settings}.
\newblock In \emph{Findings of the Association for Computational Linguistics: {ACL} 2023, Toronto, Canada, July 9-14, 2023}, pages 307--318. Association for Computational Linguistics.

\bibitem[{Venkit et~al.(2023)Venkit, Gautam, Panchanadikar, Huang, and Wilson}]{DBLP:conf/eacl/VenkitGPHW23}
Pranav~Narayanan Venkit, Sanjana Gautam, Ruchi Panchanadikar, Ting{-}Hao~K. Huang, and Shomir Wilson. 2023.
\newblock \href {https://doi.org/10.18653/V1/2023.EACL-MAIN.9} {Nationality bias in text generation}.
\newblock In \emph{Proceedings of the 17th Conference of the European Chapter of the Association for Computational Linguistics, {EACL} 2023, Dubrovnik, Croatia, May 2-6, 2023}, pages 116--122. Association for Computational Linguistics.

\bibitem[{Vig et~al.(2020)Vig, Gehrmann, Belinkov, Qian, Nevo, Singer, and Shieber}]{DBLP:journals/corr/abs-2004-12265}
Jesse Vig, Sebastian Gehrmann, Yonatan Belinkov, Sharon Qian, Daniel Nevo, Yaron Singer, and Stuart~M. Shieber. 2020.
\newblock \href {https://arxiv.org/abs/2004.12265} {Causal mediation analysis for interpreting neural {NLP:} the case of gender bias}.
\newblock \emph{CoRR}, abs/2004.12265.

\bibitem[{Wan et~al.(2023)Wan, Pu, Sun, Garimella, Chang, and Peng}]{DBLP:conf/emnlp/WanPSGCP23}
Yixin Wan, George Pu, Jiao Sun, Aparna Garimella, Kai{-}Wei Chang, and Nanyun Peng. 2023.
\newblock \href {https://aclanthology.org/2023.findings-emnlp.243} {"kelly is a warm person, joseph is a role model": Gender biases in llm-generated reference letters}.
\newblock In \emph{Findings of the Association for Computational Linguistics: {EMNLP} 2023, Singapore, December 6-10, 2023}, pages 3730--3748. Association for Computational Linguistics.

\bibitem[{Wang et~al.(2024{\natexlab{a}})Wang, Li, Zhang, Xu, Yao, Jiang, Xie, Huang, and Chen}]{wise}
Peng Wang, Zexi Li, Ningyu Zhang, Ziwen Xu, Yunzhi Yao, Yong Jiang, Pengjun Xie, Fei Huang, and Huajun Chen. 2024{\natexlab{a}}.
\newblock \href {https://arxiv.org/abs/2405.14768} {Wise: Rethinking the knowledge memory for lifelong model editing of large language models}.
\newblock \emph{CoRR}, abs/2405.14768.

\bibitem[{Wang et~al.(2024{\natexlab{b}})Wang, Wu, He, and Chen}]{washing}
Yu~Wang, Ruihan Wu, Zexue He, and Xiusi Chen. 2024{\natexlab{b}}.
\newblock \href {https://doi.org/10.48550/ARXIV.2405.14768} {Large scale knowledge washing}.
\newblock \emph{CoRR}, abs/2405.14768.

\bibitem[{Wei et~al.(2023)Wei, Yu, Ma, Lei, Weng, Song, and Liu}]{DBLP:journals/corr/abs-2311-09053}
Yifan Wei, Xiaoyan Yu, Huanhuan Ma, Fangyu Lei, Yixuan Weng, Ran Song, and Kang Liu. 2023.
\newblock \href {https://doi.org/10.48550/ARXIV.2311.09053} {Assessing knowledge editing in language models via relation perspective}.
\newblock \emph{CoRR}, abs/2311.09053.

\bibitem[{Wolf et~al.(2019)Wolf, Debut, Sanh, Chaumond, Delangue, Moi, Cistac, Rault, Louf, Funtowicz, and Brew}]{huggingface}
Thomas Wolf, Lysandre Debut, Victor Sanh, Julien Chaumond, Clement Delangue, Anthony Moi, Pierric Cistac, Tim Rault, R{\'{e}}mi Louf, Morgan Funtowicz, and Jamie Brew. 2019.
\newblock \href {https://arxiv.org/abs/1910.03771} {Huggingface's transformers: State-of-the-art natural language processing}.
\newblock \emph{CoRR}, abs/1910.03771.

\bibitem[{Wu et~al.(2023{\natexlab{a}})Wu, Peng, Chen, Su, and Sun}]{DBLP:journals/corr/abs-2308-09954}
Suhang Wu, Minlong Peng, Yue Chen, Jinsong Su, and Mingming Sun. 2023{\natexlab{a}}.
\newblock \href {https://doi.org/10.48550/ARXIV.2308.09954} {Eva-kellm: {A} new benchmark for evaluating knowledge editing of llms}.
\newblock \emph{CoRR}, abs/2308.09954.

\bibitem[{Wu et~al.(2023{\natexlab{b}})Wu, Li, Xu, Dong, Wu, Bian, and Xiong}]{DBLP:conf/emnlp/WuLXDW0X23}
Xinwei Wu, Junzhuo Li, Minghui Xu, Weilong Dong, Shuangzhi Wu, Chao Bian, and Deyi Xiong. 2023{\natexlab{b}}.
\newblock \href {https://aclanthology.org/2023.emnlp-main.174} {{DEPN:} detecting and editing privacy neurons in pretrained language models}.
\newblock In \emph{Proceedings of the 2023 Conference on Empirical Methods in Natural Language Processing, {EMNLP} 2023, Singapore, December 6-10, 2023}, pages 2875--2886. Association for Computational Linguistics.

\bibitem[{Xie and Lukasiewicz(2023)}]{DBLP:conf/acl/XieL23}
Zhongbin Xie and Thomas Lukasiewicz. 2023.
\newblock \href {https://doi.org/10.18653/V1/2023.ACL-LONG.876} {An empirical analysis of parameter-efficient methods for debiasing pre-trained language models}.
\newblock In \emph{Proceedings of the 61st Annual Meeting of the Association for Computational Linguistics (Volume 1: Long Papers), {ACL} 2023, Toronto, Canada, July 9-14, 2023}, pages 15730--15745. Association for Computational Linguistics.

\bibitem[{Yan et~al.(2024)Yan, Wang, Li, and Zhang}]{DBLP:journals/corr/abs-2402.13462}
Jianhao Yan, Futing Wang, Yafu Li, and Yue Zhang. 2024.
\newblock \href {https://doi.org/10.48550/ARXIV.2402.1346} {Potential and challenges of model editing for social debiasing}.
\newblock \emph{CoRR}, abs/2402.13462.

\bibitem[{Yang et~al.(2023)Yang, Yu, Fung, Li, and Ji}]{DBLP:conf/aaai/YangY0LJ23}
Ke~Yang, Charles Yu, Yi~Ren Fung, Manling Li, and Heng Ji. 2023.
\newblock \href {https://doi.org/10.1609/AAAI.V37I9.26279} {{ADEPT:} {A} debiasing prompt framework}.
\newblock In \emph{Thirty-Seventh {AAAI} Conference on Artificial Intelligence, {AAAI} 2023, Thirty-Fifth Conference on Innovative Applications of Artificial Intelligence, {IAAI} 2023, Thirteenth Symposium on Educational Advances in Artificial Intelligence, {EAAI} 2023, Washington, DC, USA, February 7-14, 2023}, pages 10780--10788. {AAAI} Press.

\bibitem[{Yao et~al.(2023)Yao, Wang, Tian, Cheng, Li, Deng, Chen, and Zhang}]{modeleditingsurvey}
Yunzhi Yao, Peng Wang, Bozhong Tian, Siyuan Cheng, Zhoubo Li, Shumin Deng, Huajun Chen, and Ningyu Zhang. 2023.
\newblock \href {https://aclanthology.org/2023.emnlp-main.632} {Editing large language models: Problems, methods, and opportunities}.
\newblock In \emph{Proceedings of the 2023 Conference on Empirical Methods in Natural Language Processing, {EMNLP} 2023, Singapore, December 6-10, 2023}, pages 10222--10240. Association for Computational Linguistics.

\bibitem[{Yin et~al.(2023)Yin, Jiang, Yang, and Wan}]{DBLP:journals/corr/abs-2312-05497}
Xunjian Yin, Jin Jiang, Liming Yang, and Xiaojun Wan. 2023.
\newblock \href {https://doi.org/10.48550/ARXIV.2312.05497} {History matters: Temporal knowledge editing in large language model}.
\newblock \emph{CoRR}, abs/2312.05497.

\bibitem[{Yu et~al.(2023)Yu, Jeoung, Kasi, Yu, and Ji}]{PCGU}
Charles Yu, Sullam Jeoung, Anish Kasi, Pengfei Yu, and Heng Ji. 2023.
\newblock \href {https://doi.org/10.18653/V1/2023.FINDINGS-ACL.375} {Unlearning bias in language models by partitioning gradients}.
\newblock In \emph{Findings of the Association for Computational Linguistics: {ACL} 2023, Toronto, Canada, July 9-14, 2023}, pages 6032--6048. Association for Computational Linguistics.

\bibitem[{Zhang et~al.(2024)Zhang, Yao, Tian, Wang, Deng, Wang, Xi, Mao, Zhang, Ni, Cheng, Xu, Xu, Gu, Jiang, Xie, Huang, Liang, Zhang, Zhu, Zhou, and Chen}]{knowedit}
Ningyu Zhang, Yunzhi Yao, Bozhong Tian, Peng Wang, Shumin Deng, Mengru Wang, Zekun Xi, Shengyu Mao, Jintian Zhang, Yuansheng Ni, Siyuan Cheng, Ziwen Xu, Xin Xu, Jia{-}Chen Gu, Yong Jiang, Pengjun Xie, Fei Huang, Lei Liang, Zhiqiang Zhang, Xiaowei Zhu, Jun Zhou, and Huajun Chen. 2024.
\newblock \href {https://doi.org/10.48550/ARXIV.2401.01286} {A comprehensive study of knowledge editing for large language models}.
\newblock \emph{CoRR}, abs/2401.01286.

\bibitem[{Zhao et~al.(2020)Zhao, Mukherjee, Hosseini, Chang, and Awadallah}]{DBLP:conf/acl/ZhaoMHCA20}
Jieyu Zhao, Subhabrata Mukherjee, Saghar Hosseini, Kai{-}Wei Chang, and Ahmed~Hassan Awadallah. 2020.
\newblock \href {https://doi.org/10.18653/v1/2020.acl-main.260} {Gender bias in multilingual embeddings and cross-lingual transfer}.
\newblock In \emph{Proceedings of the 58th Annual Meeting of the Association for Computational Linguistics, {ACL} 2020, Online, July 5-10, 2020}, pages 2896--2907. Association for Computational Linguistics.

\bibitem[{Zhu et~al.(2020)Zhu, Rawat, Zaheer, Bhojanapalli, Li, Yu, and Kumar}]{DBLP:journals/corr/abs-2012-00363}
Chen Zhu, Ankit~Singh Rawat, Manzil Zaheer, Srinadh Bhojanapalli, Daliang Li, Felix~X. Yu, and Sanjiv Kumar. 2020.
\newblock \href {https://arxiv.org/abs/2012.00363} {Modifying memories in transformer models}.
\newblock \emph{CoRR}, abs/2012.00363.

\bibitem[{Zmigrod et~al.(2019)Zmigrod, Mielke, Wallach, and Cotterell}]{CDA}
Ran Zmigrod, S.~J. Mielke, Hanna~M. Wallach, and Ryan Cotterell. 2019.
\newblock \href {https://doi.org/10.18653/v1/p19-1161} {Counterfactual data augmentation for mitigating gender stereotypes in languages with rich morphology}.
\newblock In \emph{Proceedings of the 57th Conference of the Association for Computational Linguistics, {ACL} 2019, Florence, Italy, July 28- August 2, 2019, Volume 1: Long Papers}, pages 1651--1661. Association for Computational Linguistics.

\end{thebibliography}

\appendix

\newpage
\section{Bias Tracing}
\label{app:tracing}
Some works \citep{DBLP:journals/corr/abs-2501-16496, lin2025survey} use causal tracing to mechanistic interpretability for LLMs.
ROME \citep{ROME} and MEMIT \citep{MEMIT} utilize causal tracing \cite{DBLP:journals/corr/abs-2004-12265} to locate facts memorized causal LMs.
After they find the specific hidden state with the strongest effect on individual facts, they modify these localized parameters for changing facts.
Inspired by causal tracing, we propose bias tracing to seek the exact hidden states that contribute most to bias exhibited in the language models including masked language models and causal language models, which will guide us to select positions to edit for debiasing.

\subsection{Tracing Bias Associations}
Following \citet{ROME}, we analyze all internal activations of a language model $\mathcal{M}$ during three runs: a clean run eliciting the bias in language models, a corrupted run disrupting the bias context modeling, and a corrupted-with-restoration run measuring bias exhibited in every single state.
\begin{itemize}
     \item As for the \textbf{clean} run, we obtain $P_\theta(x_\text{stereo})$ and $P_\theta(x_\text{anti})$ for each sample in the datasets, and collect all hidden activations $h_i^\ell$ for each token $i$ and each layer $\ell$, given the input text $x=[x_1, \ldots, x_K]$ and the $\mathcal{M}$ with $L$ layers. 
    
    \item In the \textbf{corrupted} run, noise is added to the embedding of bias attribute words in the input. For the embedding $h_i^0$ in the token sequences of bias attributes words to be corrupted, we set $\hat{h_i^0}:=h_i^0 + \tau$, where $\tau \sim \mathcal{N}(0;\sigma).$\footnote{$\sigma$ is three times the standard deviation of embeddings of 1000 subjects from \url{https://rome.baulab.info/data/dsets/known_1000.json} as \citet{ROME}} Then, $\mathcal{M}$ runs based on the corrupted embeddings and we collected the following corrupted activations $\hat{h_{i}^\ell}$.
    Since the existence of bias attribute words in a context is the reason why a context presents bias, corrupting the embedding of bias attribute words will remove the bias associations on the following language modeling process.
   
    \item With noisy embeddings, in the \textbf{corrupted-with-restoration} run, we restore specific hidden states of some token $i, i\in [0, K]$ (the bias attribute words, the attribute term, or the token before the attribute term) in an input context and layer $\ell, \ell \in [0, L]$ (the Transformer block, the attention layer, or the MLP layer) of a language model, which lets $\mathcal{M}$ output the clean state $h_i^\ell$. The following forward-running executes without more intervention. 
\end{itemize}
We calculate the absolute log probability difference between $x_\text{stereo}$ and $x_\text{anti}$, $f_d(\theta, x_\text{stereo}, x_\text{anti})=|\log P_\theta(x_{\text{stereo}}) - \log P_\theta(x_{\text{anti}})|$ , to measure bias in a language model. 
The larger the difference is, the more biased $\mathcal{M}$ is.
By running the network twice, bias tracing computes the bias association of activations.
The clean run occurs first to obtain all clean activations. 
Secondly, embeddings of bias attribute words are corrupted and the lowest difference is obtained.
Then the corrupted activations $\hat{h_i^\ell}$ of a certain token $i$ and layer $\ell$ are restored to their original values $h_i^\ell$ from the same token $i$ at the same layer $\ell$.
All differences are recorded after restoring activations over every token in the input context and every layer.
If an activation restoration of a token $i'$ and layer $\ell'$ causes a larger difference than a restoration from other tokens and layers, we can know that the activations of the token $i'$ and layer $\ell'$ give more impetus to bias.

\begin{figure*}
    \centering    
    \includegraphics[width=1\textwidth]{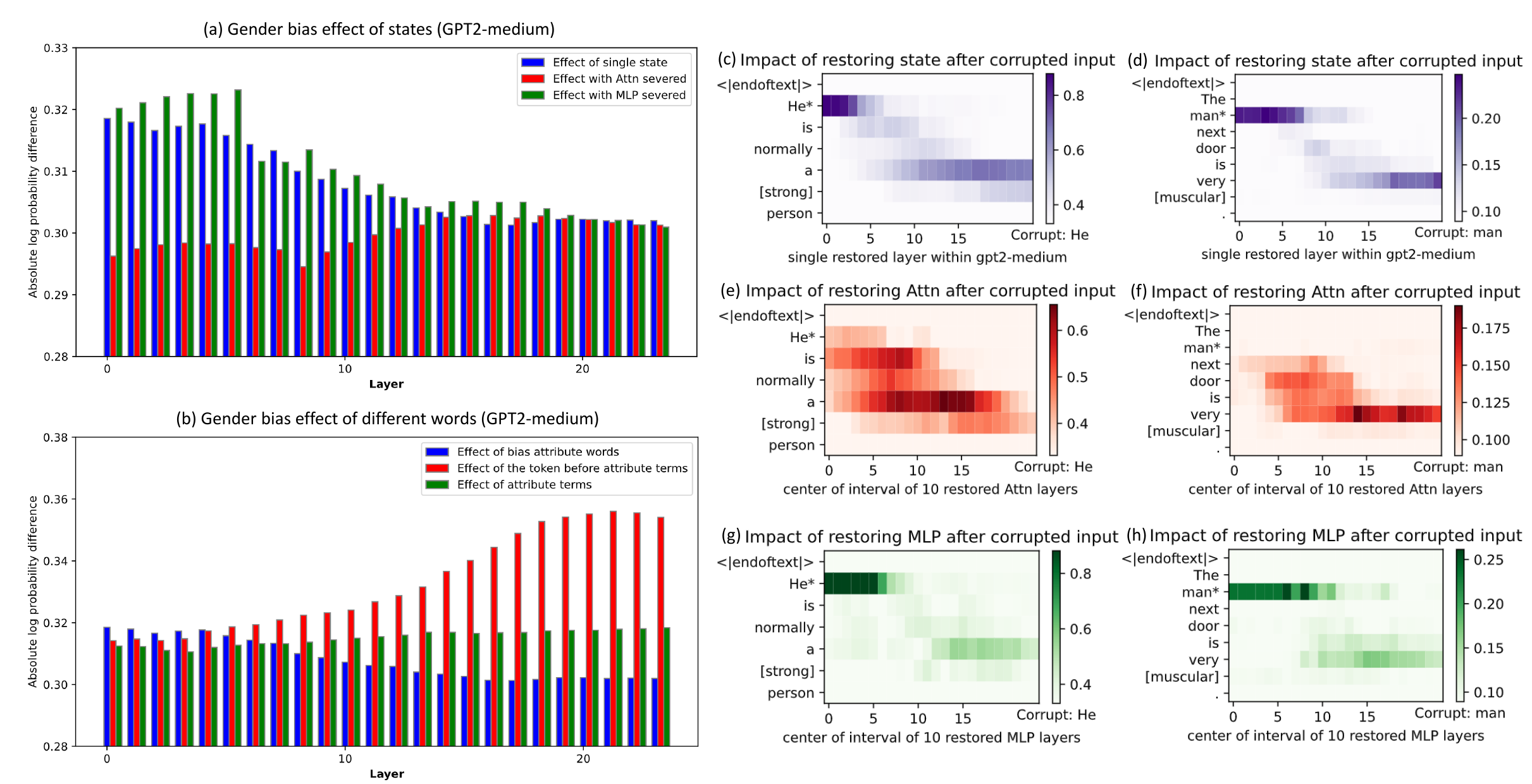}
    \caption{Gender bias tracing on GPT2-medium. (a) Comparing bias associations of bias attribute words on hidden states, attention layers, and MLP layers. (b) Comparing bias associations on single states of the bias attribute word, the token before the attribute term, and the attribute term. The bias impacts on output probability are mapped for the effect of (c-d) each hidden state on the context, (e-f) only MLP activations, and (g-h) only attention activations. * marks the corrupted bias attribute words and [] refers to the attribute terms in (c-h).}
    \label{fig:gpt2m_gender_tracing}
\end{figure*}
\begin{figure*}
    \centering    
    \includegraphics[width=1\textwidth]{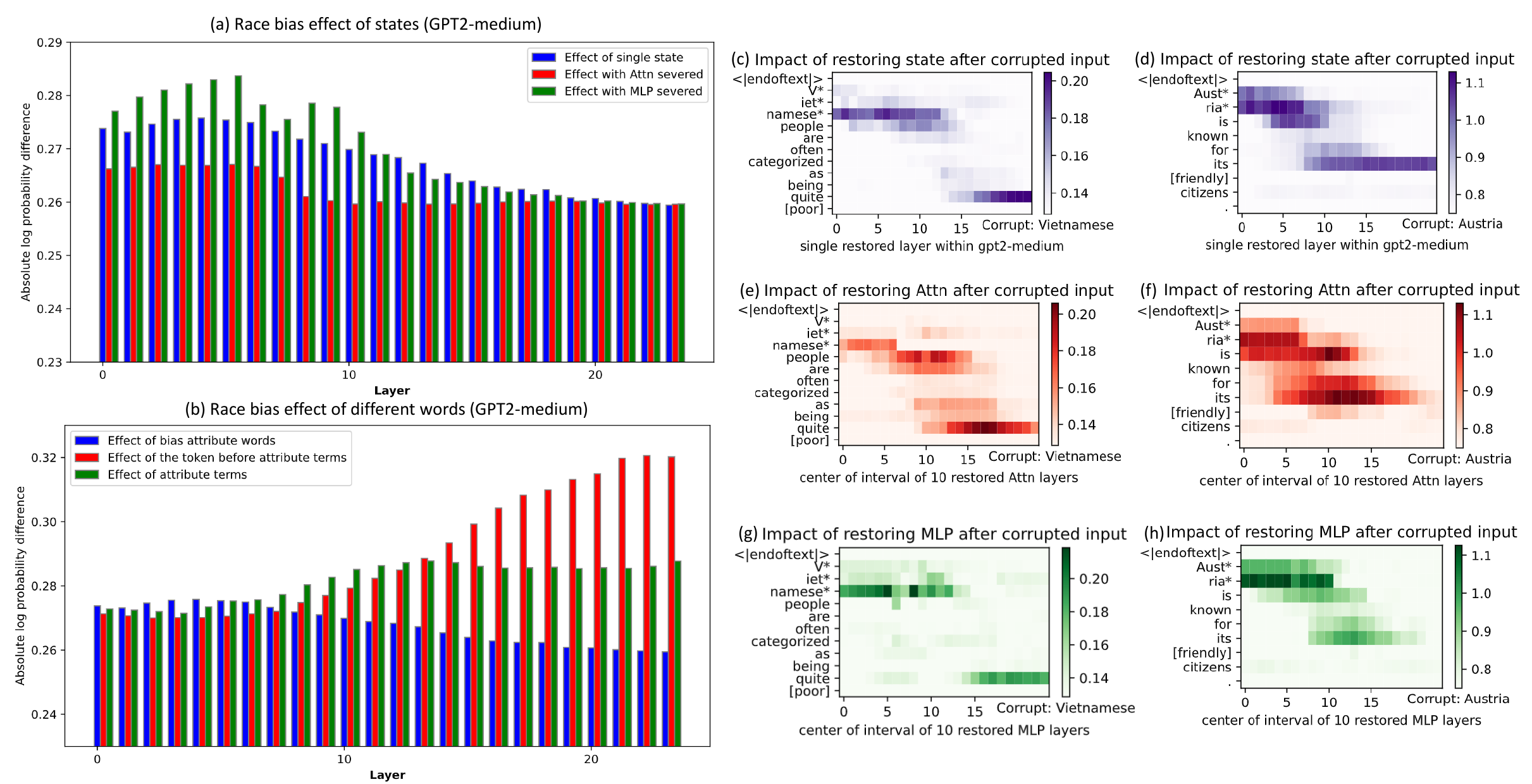}
    \caption{Race bias tracing on GPT2-medium.}
    \label{fig:gpt2m_race_tracing}
\end{figure*}

\subsection{Tracing Data Construction}
\label{app:tdc}
We conduct gender and race bias tracing in this paper. Therefore, gender and race bias attribute words are extracted in the context.
We begin with utilizing SPARQL to query the instance of gender and race in Wikidata, obtaining a variety of words targeted to specific bias.
These words are the source collection of bias attribute words.
Based on the collection, we then adopt simple string matching to extract bias attribute words from the context sentence $x$ of each sample $s$ in the dataset.
As a result, we can trace the activations of these bias attribute words in language models.

\subsection{Bias Tracing with GPT2}
We conduct gender and race bias tracing on the \textit{intrasentence} part of StereoSet at every layer of language models and every token in contexts.
The average bias associations of 500 samples with GPT2-medium are shown in Figure \ref{fig:gpt2m_gender_tracing} and \ref{fig:gpt2m_race_tracing}.

\paragraph{Bias best corresponds to the states of MLPs at lower layers.} Figure \ref{fig:gpt2m_gender_tracing} (a) illustrates that at layer 0-5 (layer 0-10 in Figure \ref{fig:gpt2m_race_tracing}), MLPs in transformer blocks play a much more significant role in bias than attention layers, with peaking at layer 5 while bias associations of attention layers varies a little among different blocks.
This reveals that language models intensively present bias in the foundational representations learned by lower layers, and these early presentations can influence the subsequent layers.
The reason is that since the lower layers capture the text patterns \cite{DBLP:conf/emnlp/GevaSBL21}, bias patterns in the pre-trained corpus, such as bias attribute words' cooccurrence with stereotyped terms, are memorized in the early layers. 
Figure \ref{fig:gpt2m_gender_tracing}(b) and \ref{fig:gpt2m_race_tracing}(b) also show that bias attribute words have the most effects at the early layers. 
Meanwhile, it indicates that the token before attribute terms associates a lot with bias at the upper layers of causal language models because semantic information is usually modeled in the top layers and the attribute term explicitly semantically presents bias.
Two cases in Figure \ref{fig:gpt2m_gender_tracing}(c-h) and \ref{fig:gpt2m_race_tracing}(c-h) illustrate the aforementioned observations well.



\section{Experimental Details}

\label{app:appexp}

\subsection{StereoSet}
\label{app:stereoset}
\begin{table}[htbp]
  \centering
    \scalebox{0.99}{
    \begin{tabular}{lccc}
    \toprule
      & \textbf{\# Gender} & \textbf{\# Race} & \textbf{\# Religion} \\
    \midrule
    \textbf{$\mathcal{S}_\text{edit}^\text{train}$} & 617 & 2,307 & 210 \\
    \textbf{$\mathcal{S}_\text{edit}^\text{dev}$} & 70 & 297 & 25 \\
    \textbf{$\mathcal{S}_\text{edit}^\text{test}$} & 253 & 962 & 77 \\
    \bottomrule
    \end{tabular}}
    \caption{The numbers of samples about different bias in our dataset.}
     \label{tab:stereoset}
\end{table}

\begin{table*}[htbp]
  \centering
  \scalebox{0.98}{
    \begin{tabular}{lcccccc}
    \toprule
    \multicolumn{1}{c}{\multirow{2}[4]{*}{Method}} & \multicolumn{3}{c}{\textbf{GPT2-medium}} & \multicolumn{3}{c}{\textbf{Gemma-2b}} \\
\cmidrule{2-7}          & Gender & Race  & Religion & Gender & Race  & Religion \\
    \midrule
    \midrule
    \textbf{Pre-edit} & 61.46  & 59.57  & 73.33  & 63.54  & 64.54  & 66.67  \\
    \midrule
    CDA   & 51.04  & 44.68  & 66.67  & \multicolumn{3}{c}{\textbf{-}} \\
    SentenceDebias & 56.33 & 55.48 & 53.14 & 60.42 & 60.99 & 61.29 \\
    Self-Debias & \textbf{50.00} & 59.57 & 53.33 & 56.25 & 43.26 & 56.25 \\
    INLP  & 47.92  & 52.81  & 61.29  & 63.57  & 60.99  & 63.33  \\
    \textbf{EditBias} & 53.08 & \textbf{50.35 } & \textbf{53.12} & \textbf{52.81} & \textbf{49.83} & \textbf{53.17} \\
    \midrule
    \multicolumn{1}{c}{\multirow{2}[4]{*}{Method}} & \multicolumn{3}{c}{\textbf{Mistral-7B-v0.3}} & \multicolumn{3}{c}{\textbf{Llama3-8B}} \\
\cmidrule{2-7}          & Gender & Race  & Religion & Gender & Race  & Religion \\
    \midrule
    \midrule
    \textbf{Pre-edit} & 65.62  & 68.09  & 70.00  & 62.50  & 62.41  & 73.33  \\
    \midrule
    CDA   & \multicolumn{6}{c}{\textbf{-}} \\
    SentenceDebias & 61.46 & 66.67 & 70.00 & 60.42 & 61.49 & 62.50 \\
    Self-Debias & 41.67 & 41.89  & 40.00  & 44.79  & 47.52  & \textbf{46.67 } \\
    INLP  & 59.38 & 68.79 & 68.75  & 56.25  & 63.83  & 70.00  \\
    \textbf{EditBias} & \textbf{49.65} & \textbf{48.94} & \textbf{53.24} & \textbf{52.39} & \textbf{50.17} & 54.94 \\
    \bottomrule
    \end{tabular}}
\caption{Stereotype Score (\%) for evaluating the baselines and \ours\ on Crows-Pairs.}
  \label{tab:crowspairs}
\end{table*}

\begin{table*}[htbp]
  \centering
  \scalebox{0.88}{
    \begin{tabular}{lcccccccc}
    \toprule
    \multicolumn{1}{c}{\multirow{3}[4]{*}{Bias\newline{}Type}} & \multicolumn{4}{c}{\textbf{GPT2-medium}} & \multicolumn{4}{c}{\textbf{Gemma-2b}} \\
\cmidrule{2-9}          & \multicolumn{2}{c}{\textbf{One}} & \multicolumn{2}{c}{\textbf{Mixture}} & \multicolumn{2}{c}{\textbf{One}} & \multicolumn{2}{c}{\textbf{Mixture}} \\
          & \textbf{SS (\%)} & \textbf{$\Delta$LMS (\%)} & \textbf{SS (\%)} & \textbf{$\Delta$LMS (\%)} & \textbf{SS (\%)} & \textbf{$\Delta$LMS (\%)} & \textbf{SS (\%)} & \textbf{$\Delta$LMS (\%)} \\
    \midrule
    \midrule
    Gender & 49.81 & -1.22 & 49.42 & -8.82 & 47.71 & -5.36 & 48.59 & -4.78 \\
    Race  & 55.27 & -5.57 & 56.34 & -5.12 & 54.88 & -2.39 & 55.86 & -4.35 \\
    Religion & 49.64 & -6.94 & 53.55 & -1.92 & 50.42 & -8.53 & 47.36 & -5.44 \\
    \midrule
    \multicolumn{1}{c}{\multirow{3}[4]{*}{Bias\newline{}Type}} & \multicolumn{4}{c}{\textbf{Mistral-7B-v0.3}} & \multicolumn{4}{c}{\textbf{Llama3-8B}} \\
\cmidrule{2-9}          & \multicolumn{2}{c}{\textbf{One}} & \multicolumn{2}{c}{\textbf{Mixture}} & \multicolumn{2}{c}{\textbf{One}} & \multicolumn{2}{c}{\textbf{Mixture}} \\
          & \textbf{SS (\%)} & \textbf{$\Delta$LMS (\%)} & \textbf{SS (\%)} & \textbf{$\Delta$LMS (\%)} & \textbf{SS (\%)} & \textbf{$\Delta$LMS (\%)} & \textbf{SS (\%)} & \textbf{$\Delta$LMS (\%)} \\
    \midrule
    \midrule
    Gender & 48.96  & -10.55 & 46.24  & -8.81 & 50.00  & -10.98 & 49.18  & -13.42 \\
    Race  & 53.32  & -6.25 & 51.46  & -8.59 & 46.28  & -20.84 & 53.51  & -11.77 \\
    Religion & 52.15  & -7.72 & 50.42  & -0.03 & 50.42  & -8.56 & 51.13  & -10.02 \\
    \bottomrule
    \end{tabular}}
    \caption{Training editor networks with data for one type of bias vs. mixed types of bias.}
  \label{tab:onetype}
\end{table*}

\subsection{Settings}
We use four pre-trained language models in our experiments from HuggingFace \cite{huggingface}, including GPT2-medium\footnote{\url{https://huggingface.co/openai-community/gpt2-medium}}, Gemma-2B\footnote{\url{https://huggingface.co/google/gemma-2b}}, Mistral-7B-v0.3\footnote{\url{https://huggingface.co/mistralai/Mistral-7B-v0.3}}, and Llama3-8B\footnote{\url{https://huggingface.co/meta-llama/Meta-Llama-3-8B}}.
For each training, we use one A800 80GB GPU and grid search among [8, 16, 64] batch sizes for batch editing.
The $\lambda$ is determined by grid searching in \{1.0, 2.0, 3.0, 4.0, 5.0\}.

\subsection{Baselines}
\label{app:baselines}
\paragraph{CDA (Counterfactual Data Augmentation) \cite{CDA, barikeri-etal-2021-redditbias}} retrains a pre-trained language model.
It generates and incorporates data representing what could have happened under different conditions. 
By altering aspects of data related to biased attributes, such as changing gender or race in a dataset, a counterfactual data set is created to create a more balanced training environment for models.

\paragraph{SentenceDebias \cite{sentencedebias}} first estimates the demographic bias subspace 
by encoding sentences containing bias attribute words or their counterfactuals into sentence representations and using principle component analysis \cite{PCA} to define the bias subspace as the first K principle components, and then debiases sentence representations by subtracting their projection onto the bias subspace.

\paragraph{Self-Debias \cite{self-debias}} first prompts a model to generate toxic text, such as encouraging a model to discriminate based on gender.
Then, the model can generate a non-discriminative continuation, during which the probabilities of tokens that were prominent in the toxic generation are deliberately scaled down.

\paragraph{INLP \cite{INLP}} introduces Iterative Null-space Projection (INLP), a method that reduces bias in word embeddings by iteratively projecting them onto the null space of bias terms using a linear classifier. 
This method constructs a projection matrix to project input onto the null space of the linear classifier, continuously updating both the classifier and the projection matrix.

\subsection{Training for one bias type vs. a mixture of multiple bias types}
\label{app:onetype}
Our goal is to efficiently deal with various types of bias in one training.
We need to know if there is a debiasing performance drop if we don't deal with each bias type one by one.
Therefore, we try to train editor networks with samples of one bias type and samples of a mixture of three bias types, respectively.
Table \ref{tab:onetype} shows the comparison.
The results indicate that training with a mixture of bias-type data is comparable with one bias-type data, indicating \ours~'s capability to deal with multiple types of bias simultaneously.

\subsection{Evaluation on Crows-Pairs}
We also use Crows-Pairs \cite{crows-pairs} to evaluate the debiasing generality of \ours.
Crows-Pairs is a Crowdsourced Stereotype Pairs benchmark covering nine types of bias.
We use 262 gender samples, 516 race samples, and 105 religion samples.
In each sample, there are two sentences: a more stereotyped sentence and a less stereotyped one, which are regarded as $x_\text{stereo}$ and $x_\text{anti}$ respectively.
\textit{SS} for the baselines and \ours on Crows-Pairs are shown in Table \ref{tab:crowspairs}.

\section{Gender Counterfactual Test Set}
\label{app:reverse}
We utilize the method mentioned in Appendix \ref{app:tdc} to extract gender attribute words in gender bias samples.
These gender attribute words are reversed into their counterfacts.
Then the labels ``stereotype'' and ``anti-stereotype'' are exchanged for each sentence.
For instance, after reverse, the stereotyped context in Figure \ref{fig:method} is ``Boys tend to be more determined than girls.'' and the anti-stereotyped context is ``Boys tend to be more soft than girls.''.





\end{document}